\documentclass[10pt,twocolumn,letterpaper]{article}

\usepackage{cvpr}
\usepackage{times}
\usepackage{epsfig}
\usepackage{graphicx}
\usepackage{amsmath}
\usepackage{amssymb}
\usepackage{multirow}
\usepackage[caption=false]{subfig}
\usepackage{microtype}      
\usepackage[american]{babel}

\graphicspath{{figure/}}

\usepackage[pagebackref=true,breaklinks=true,letterpaper=true,colorlinks,bookmarks=false]{hyperref}

\cvprfinalcopy 

\def\cvprPaperID{3381} 

\ifcvprfinal\fi
\begin{document}

\title{RPC: A Large-Scale Retail Product Checkout Dataset}

\author{Xiu-Shen Wei$^1$\thanks{X.-S. Wei is the corresponding author ({\tt weixs.gm@gmail.com}).}, Quan Cui$^{1,2}$\thanks{Q. Cui's contribution was made when he was an intern in Megvii Research Nanjing.}, Lei Yang$^1$, Peng Wang$^3$, and Lingqiao Liu$^3$\\
$^1$Megvii Research Nanjing, Megvii Technology Ltd., Nanjing, China\\
$^2$Graduate School of IPS, Waseda University, Fukuoka, Japan\\
$^3$ School of Computer Science, The University of Adelaide, Adelaide, Australia\\
}

\maketitle


\begin{abstract}
Over recent years, emerging interest has occurred in integrating computer vision technology into the retail industry. Automatic checkout (ACO) is one of the critical problems in this area which aims to automatically generate the shopping list from the images of the products to purchase. The main challenge of this problem comes from the large scale and the fine-grained nature of the product categories as well as the difficulty for collecting training images that reflect the realistic checkout scenarios due to continuous update of the products. Despite its significant practical and research value, this problem is not extensively studied in the computer vision community, largely due to the lack of a high-quality dataset. To fill this gap, in this work we propose a new dataset to facilitate relevant research. Our dataset enjoys the following characteristics: (1) It is by far the largest dataset in terms of both product image quantity and product categories. (2) It includes single-product images taken in a controlled environment and multi-product images taken by the checkout system. (3) It provides different levels of annotations for the check-out images. Comparing with the existing datasets, ours is closer to the realistic setting and can derive a variety of research problems. Besides the dataset, we also benchmark the performance on this dataset with various approaches. The dataset and related resources can be found at \url{https://rpc-dataset.github.io/}.
\end{abstract}


\section{Introduction}
\begin{figure}
\centering
	{\includegraphics[width=\columnwidth]{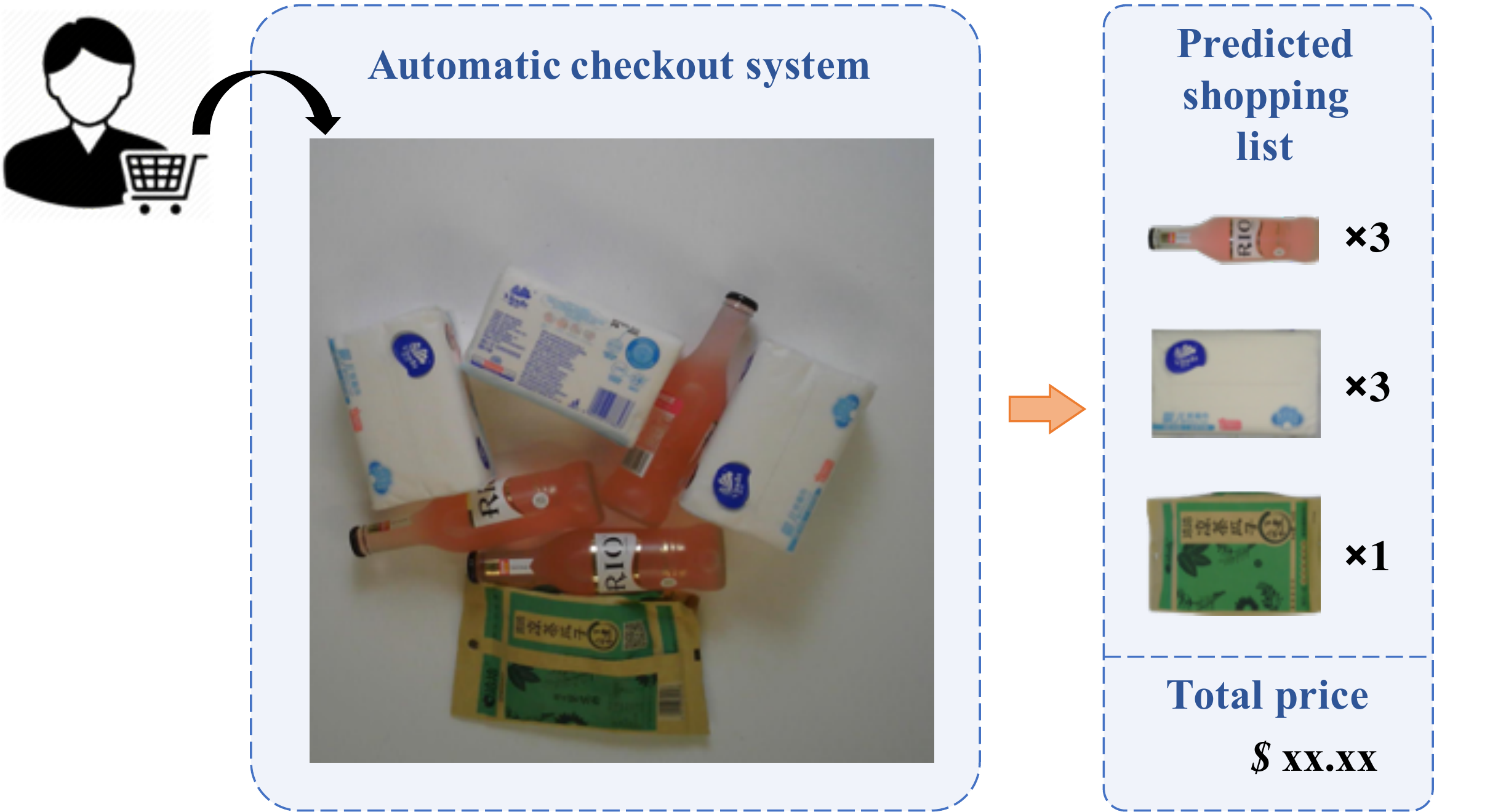}}
\caption{Illustration of the automatic checkout (ACO) application scenario. When a customer puts his/her collected products on the checkout counter, the system will automatically recognize each product and returns a complete shopping list with total price.}
\label{fig:checkoutprocess}
\end{figure}

The retail industry requires a huge amount of human labor and a large percentage of the workload is spent on recognizing products. With the recent development of computer vision, it becomes increasingly demanding to use image recognition technologies to automate the products recognition. As a primary user-case of this trend, automatic checkout (ACO) which aims to generate the shopping list from the images of the products to purchase receives emerging interests. From the image recognition perspective, this problem is particularly challenging: the number of products in a supermarket can be huge and the difference between similar products can be subtle; moreover, since it can be impractical to collect a large number of training images per product, the training of the recognition model has to deal with the small training sample size. Even worse, in some cases, is that we may only have access to the product images taken in an environment different from the deployment scenario and there is a substantial domain shift from the training set to the test set. Therefore, the ACO problem can have the characteristic of large-scale, fine-grained, few-shot and cross-domain. Each of those factors has been considered to be challenging in the computer vision literature. 

Despite its potential practical and research value, the ACO problem is not well studied in the computer vision community. This is largely due to the lack of a high-quality dataset with a clearly defined setting. To fill this gap, in this work we propose a new dataset to facilitate future research on this topic. The design of our dataset mimics the real-world scenarios in ACO. More specifically, it contains a large number of images and product categories. Some of the product categories are visually very similar and this reflects the fine-grained property in the ACO problem. This dataset also provides images of two different types. One type is taken in a controlled environment and only contains a single product. This can correspond to product images on the advertisement website. Another type represents images of user-purchased products and these images usually include multiple products. For the second type of images, we also provide different levels of annotations and clutter degrees. Researchers can use those annotations to define related sub-problems, such as detection or counting. Comparing with the existing datasets, our dataset is much closer to a realistic scenario, and a variety of research problems and settings can be derived from the proposed dataset. To benchmark this dataset, we also proposed several detection-based baselines. The last baseline, which trains detectors by incorporating an effective Cycle-GAN~\cite{zhu2017unpaired} based data augmentation scheme is identified as the best solution so far.
From the baseline results, it can be seen that the problem defined on the proposed dataset is challenging and leaves substantial room for improvement. 

\section{Related work}

\paragraph{Related datasets review:}
In this part, we review some of the existing datasets relevant to our task.

$\bullet$ \textbf{SOIL-47}~\cite{koubaroulis2002evaluating} is a product dataset that focuses on testing color-based object recognition algorithms. It contains 47 product categories, and for each category 21 images are captured from 20 different horizontal views. Two sets of such images are captured under different light conditions to test algorithms claiming illumination intensity invariance. 

$\bullet$ \textbf{Supermarket Produce Dataset}~\cite{rocha2010automatic} is introduced for automatic fruit and vegetable classification from images. The dataset has 15 product categories comprising 2,633 images captured under diverse conditions. The task, however, has nearly been solved as implemented solution achieves classification error under 2\% on the dataset.

$\bullet$ \textbf{Grozi-120}~\cite{merler2007recognizing} is a dataset proposed for groceries recognition in natural environment. It contains 120 grocery product categories. For each product category, two types of images are collected. While one type of images are collected from the web, the other type of images are collected inside a grocery store. In total, 11,870 images are collected with 676 from the web and 11,194 from the store. Traditional algorithms, such as color histogram matching, SIFT matching, and boosted Haar-like features, are applied to the dataset for performance evaluation.

$\bullet$ \textbf{Grocery Products Dataset}~\cite{george2014recognizing} is another dataset aiming at grocery product recognition. It contains 80 grocery products and collects 8,350 training images and 680 test images. The training images are downloaded from the web, and the test images are collected in natural shelf scenario. Normally the products are neatly placed on the shelf, which is different from the realistic checkout scenario where the products are freely placed on counter in clutter.

$\bullet$ \textbf{Freiburg Groceries Dataset}~\cite{jund2016freiburg} is a grocery dataset comprising 5,021 images of 25 grocery classes. These images are divided into two sets: a training set that consists of 4,947 images taken by smartphone cameras, each containing one or more instances of one class; a test set with 74 images of 37 clutter scenes, each containing objects of multiple classes.

$\bullet$ \textbf{MVTec D2S}~\cite{ulrichmvtec} is a dataset proposed for instance-aware semantic segmentation in an industrial domain. It provides 21,000 images of 60 object categories with pixel-wise labels. This dataset can be used as complementary to other datasets~\cite{pascal-voc-2012,mscoco,Cordts2016Cityscapes} in computer vision for grocery-relevant semantic segmentation.


\paragraph{Data augmentation in computer vision:}
Data augmentation scheme is explored in our work to mitigate the gap between training and test images. In this part, we review relevant works for data augmentation.
Naturally, if a model has a large number of parameters (\eg, deep convolutional neural networks~\cite{NIPS2012_4824,Simonyan14c,He2016DeepRL,hu2018cvprsenet}), it requires proportional amount of training samples to learn the model. Data augmentation is a common strategy used in computer vision to deal with data shortage. Conventionally, simple alternations are made on the existing dataset to expand the data size. These operations include flip, translation, scaling, rotation, random crop and so on. Apart from enabling more training data, the expansion also makes the model to be invariant to some conditions, \eg, rotation invariance, and thus obviously boosts the neural network performance~\cite{NIPS2012_4824,Simonyan14c,He2016DeepRL,hu2018cvprsenet}.

To deal with data shortage involving domain shift~\cite{da}, more advantaged data augmentation is explored and exploited. For example, the work in \cite{ulrichmvtec} synthesizes new images containing multiple objects by combining and reorganizing atom object masks. In real world, data can exist in infinite conditions. To prevent models to understand the world spuriously and superficially, data depicting various conditions is required. Generative models, \eg, Variational Auto-Encoder (VAE)~\cite{journals/corr/KingmaW13}, Generative Adversarial Networks (GANs)~\cite{GANs} and its variant~\cite{NIPS2016_6399,zhu2017unpaired,8237572,Huang_2018_ECCV}, provide a promising solution to this problem. For example, the work in~\cite{DBLP:journals/corr/abs-1709-00663} uses conditional VAE to generate visual features for novel classes to address Zero-Shot~\cite{5206594,wang2018gcnzeroshot} learning problem. In \cite{Zhou_2018_CVPR}, a method variant to GANs is proposed to generate multi-view vehicle feature based on single-view feature to boost vehicle re-identification. In some other works, GANs are directly used to generate images crossing domains or styles. For example, in~\cite{Huang_2018_ECCV}, a structure-aware image-to-image translation scheme is proposed to synthesize large-scale training data for vehicle detection. In another work~\cite{tobin2017domain}, a deep neutral network is trained on simulated images but can successfully generalize to real images by resorting to the variability in the simulator.

\section{A new dataset for the automatic checkout (ACO) problem}

\subsection{Automatic product checkout: a new computer vision task}
This subsection formally defines the problem setting for the automatic checkout (ACO) task. To begin with, we will briefly review the application scenario of the ACO problem.

When a customer puts his/her selected products on the checkout counter, an ideal ACO system is expected to be able to accurately recognize each of these products and return a complete shopping list \emph{at one glance}, as shown in Fig.~\ref{fig:checkoutprocess}. Thus the key of an ACO system is a recognition system that can accurately predict the presence and count of each product in an arbitrary product combination. 
Usually, such a recognition system is trained with the images captured at the same environment as the deployment scenario. In the context of the ACO problem, the training image should be the one taken at the checkout counter, which captures a combination of multiple product instances (we call it the checkout image hereafter). However, due to a large number of product categories as well as the continuous update of the stock list, it is infeasible to learn the recognition model by enumerating all the product combinations. In fact, it is even impractical to assume that the checkout images cover every single product on the stock list. A more economical solution is to train the recognition system by using images of each isolated product taken in a controlled environment. Once taken, those images can be reused and distributed to different deployment scenarios.


Inspired by the above scenario, we formally define the ACO problem as follows: Given a set of candidate products $\mathcal{P}=\{p_i\}$ and a test image from the test set $I_t \in \mathcal{T}$, the task is to predict the presence and count of each product in the test image, in other words, predicting $count(p)~\forall p \in \mathcal{P}$, where $count(p)$ indicates the number of occurrences in the test image and $count(p)=0$ if the product does not appear. To perform this prediction and build the model, we will have a single-product image set $\mathcal{S} = \{(I_s,y_s)|y_s \in \mathcal{P}\}$, where $I_s$ is a single-product image and $y_s$ is its associated product ID/category. Also, we may have access to a checkout image set $\mathcal{C} \{(I_c,Y_c)\}$ with optional availability of a certain level of annotations $Y_c$, \eg, $Y_c$ can correspond to the annotations in Sec.~\ref{sect:annotation}.


\begin{figure}[t]
\centering
	{\includegraphics[width=\columnwidth]{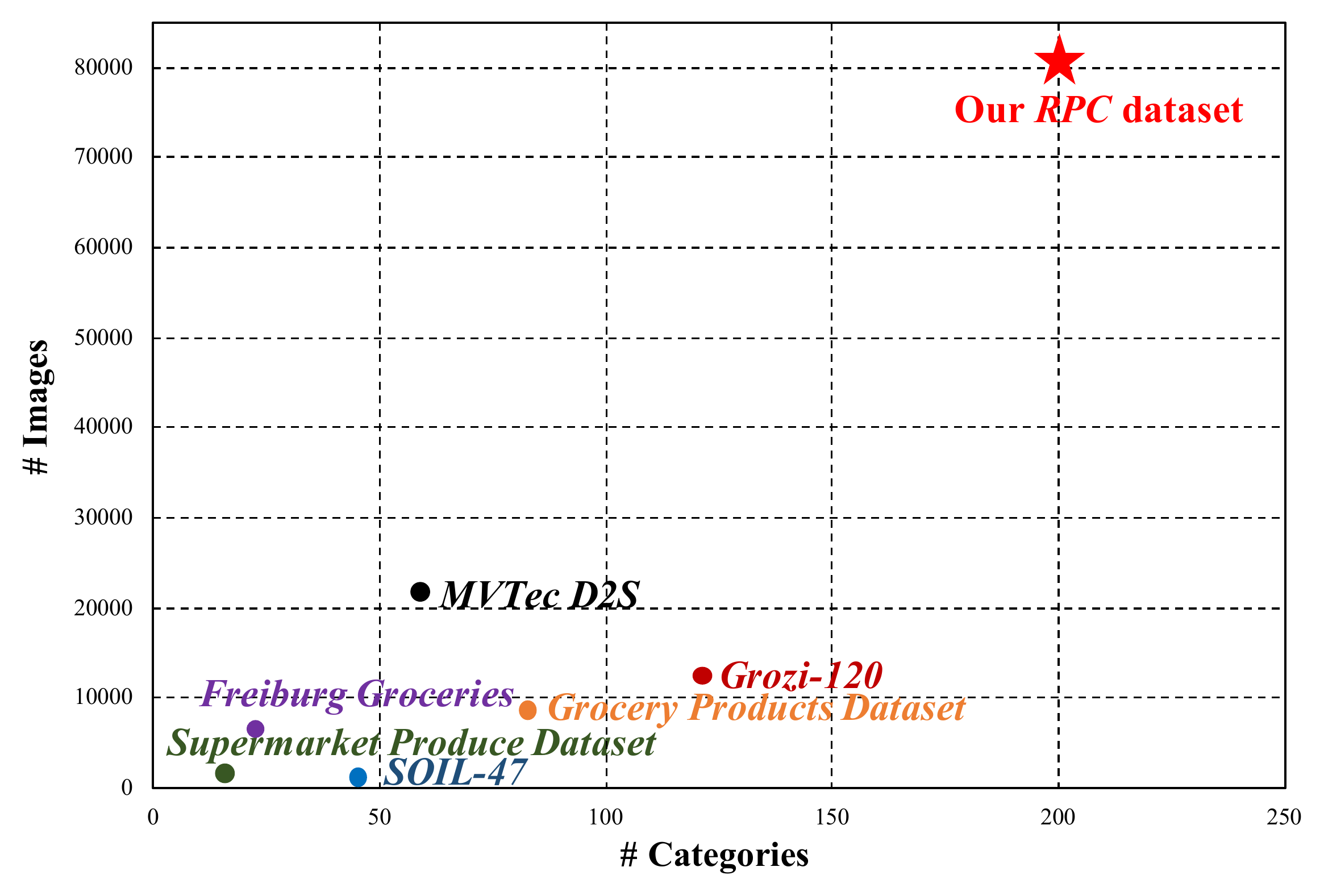}}
\caption{Comparisons with other related datasets in the literature.}
\label{fig:plot}
\end{figure}

\subsection{Proposed retail product checkout (RPC) dataset}
In this section, we elaborate the details of the proposed retail product checkout (RPC) dataset. To start with, we will introduce the characteristics of the dataset. Then, the construction details for the dataset will be given. Finally, we present the evaluation metrics adopted in this paper for the ACO problem. 

\subsubsection{Characteristics of the proposed RPC dataset}
\begin{figure}[t!]
\centering
	{\includegraphics[width=\columnwidth]{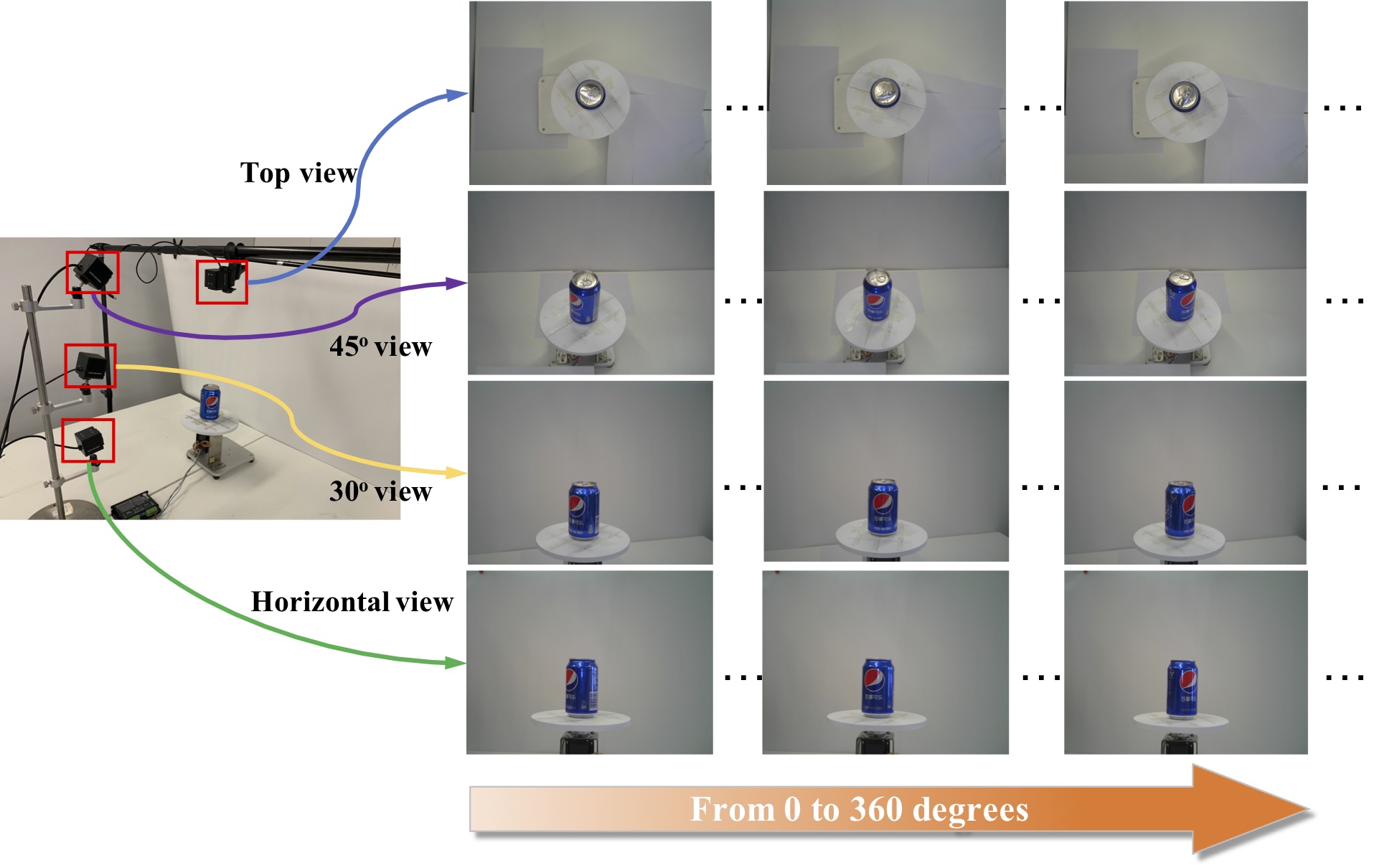}}
\caption{Collection equipment for single product images.}
\label{fig:singledesk}
\end{figure}

\begin{figure}[t!]
	\centering
	\subfloat[Easy mode.]  { \includegraphics[width=\columnwidth]{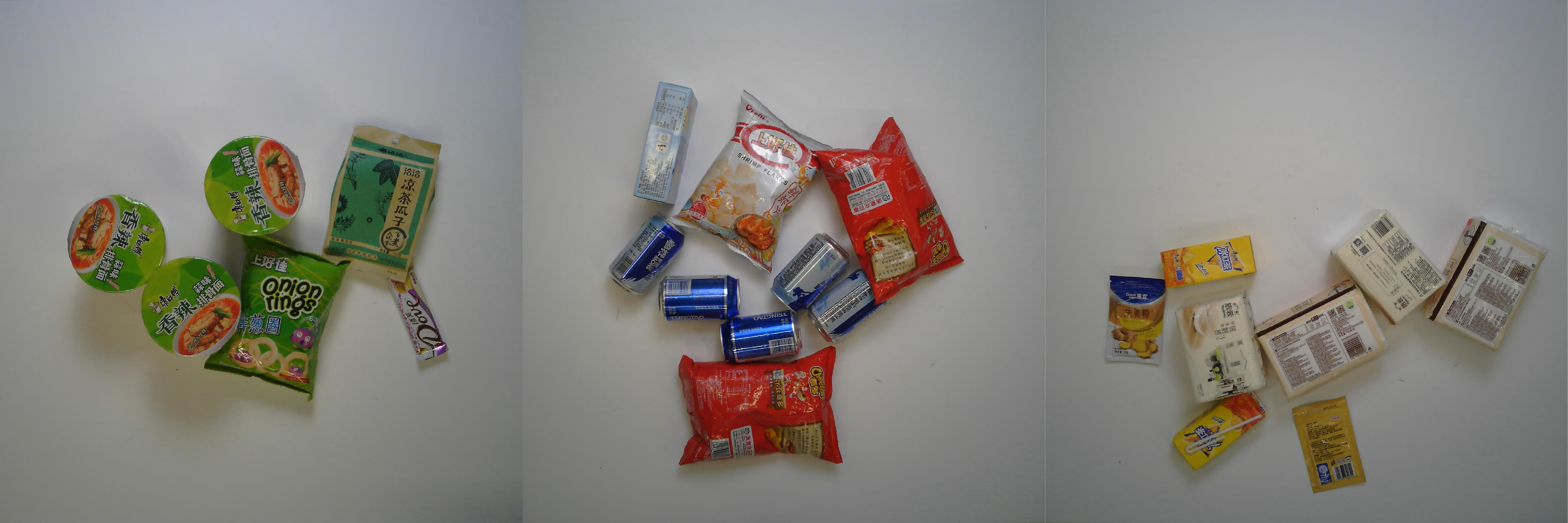} \label{fig:FG_easy_checkout} }
	
	\subfloat[Medium mode.] { \includegraphics[width=\columnwidth]{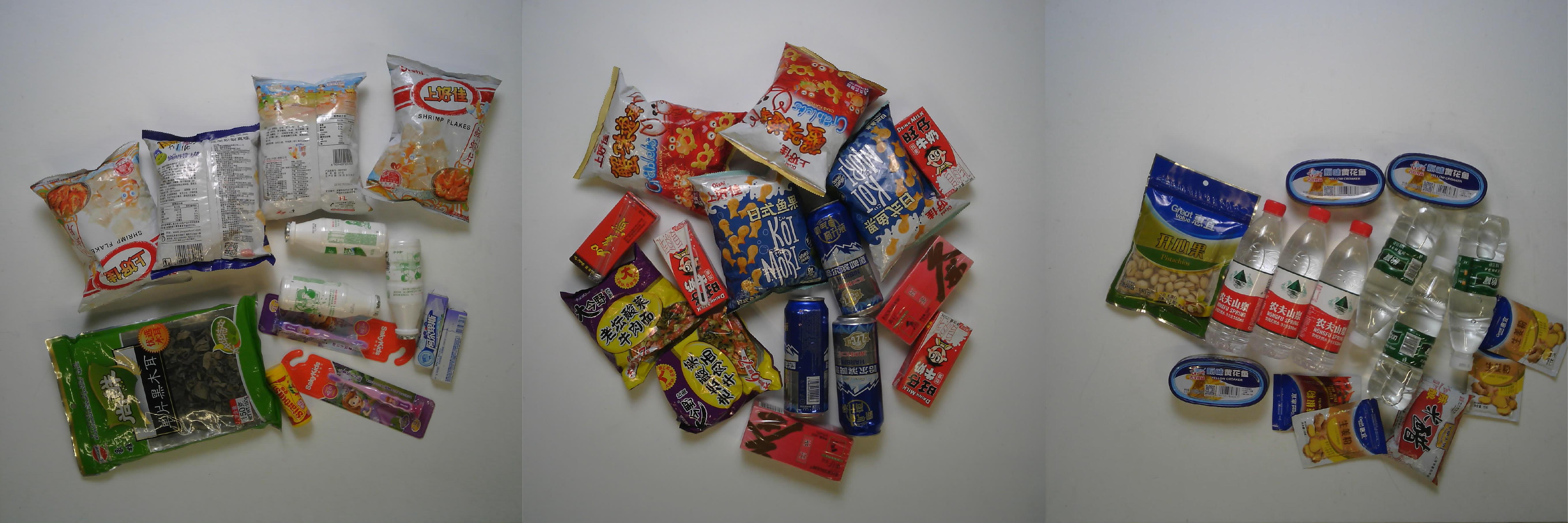} \label{fig:FG_medium_checkout} }
	
	\subfloat[Hard mode.] { \includegraphics[width=\columnwidth]{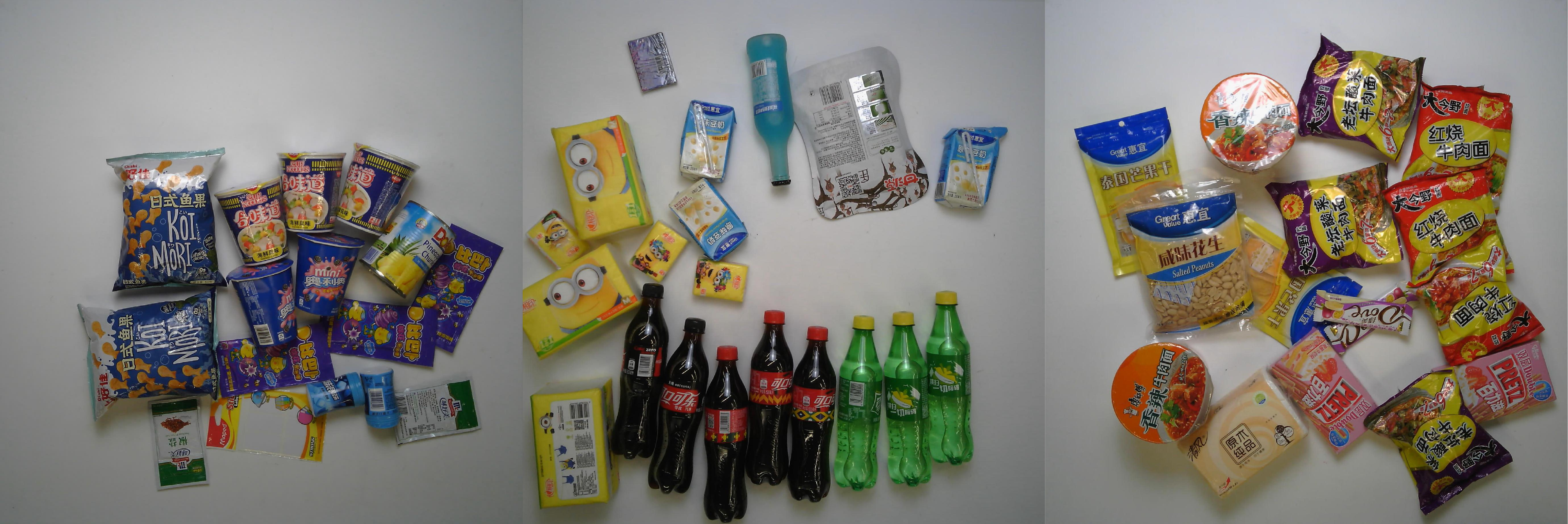} \label{fig:FG_hard_checkout} }
	\caption{Sampled checkout images of three clutter levels.} \label{fig:checkoutimgs}
\end{figure}

\begin{figure}[t!]
\centering
	{\includegraphics[width=\columnwidth]{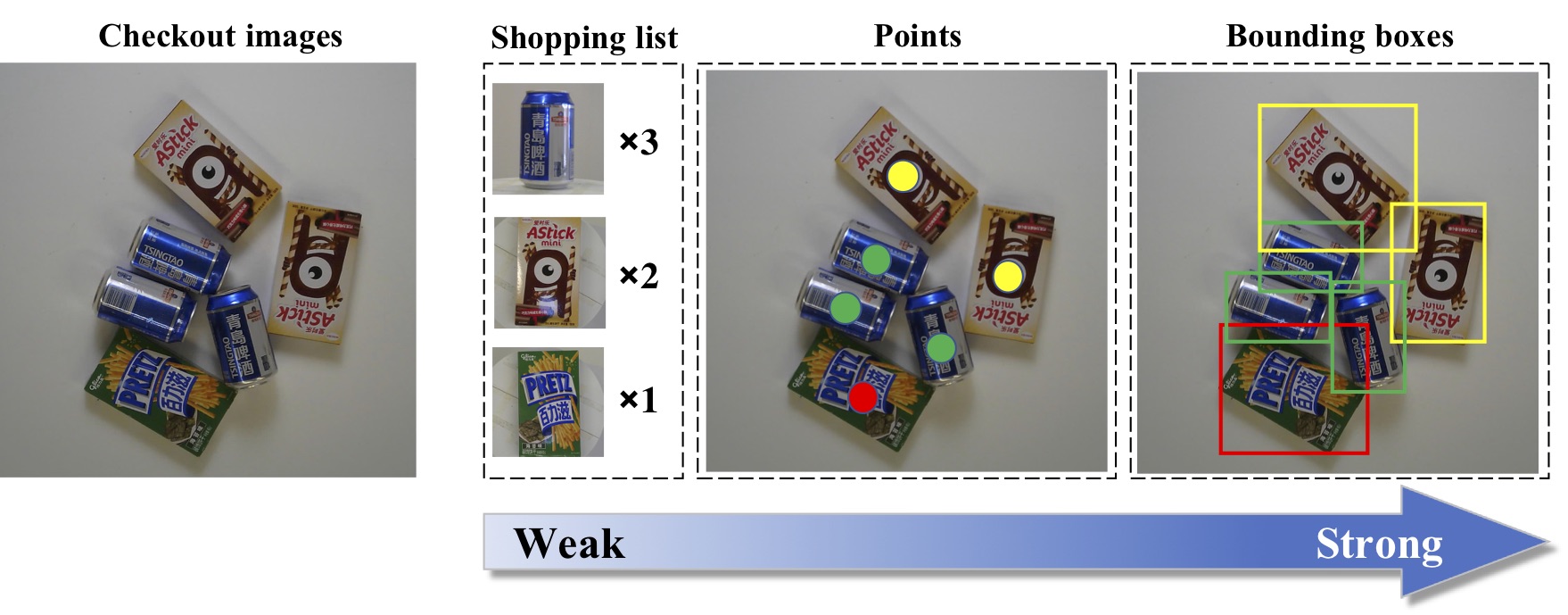}}
\caption{Weak to strong supervisions of our RPC dataset: from shopping list, points, to bounding boxes.}
\label{fig:supervision}
\end{figure}

\begin{figure*}[t!]
\centering
	{\includegraphics[width=0.99\textwidth]{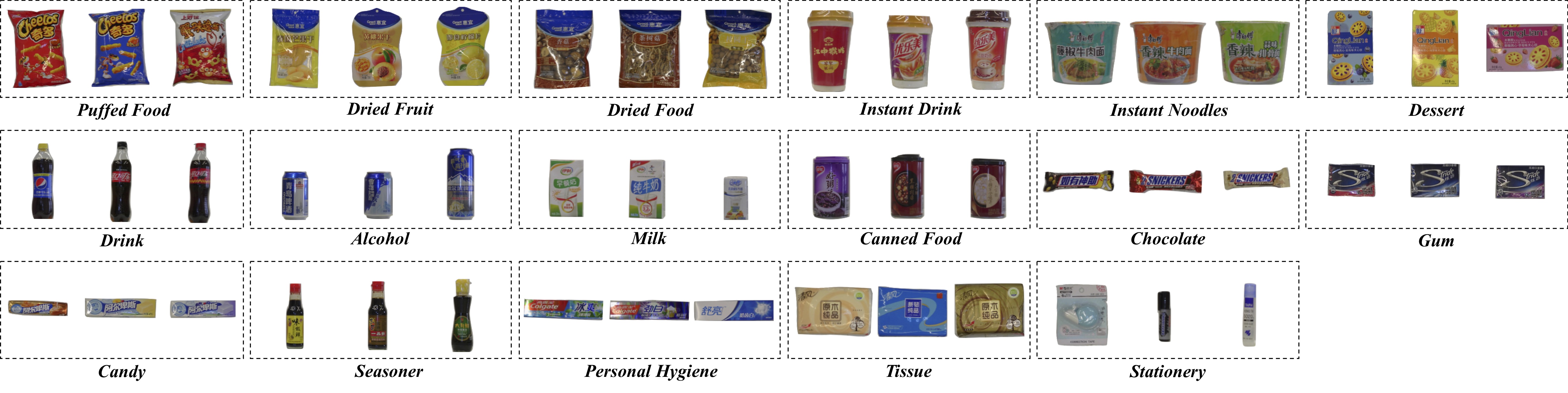}}
\caption{Sampled images of totally 200 retail products belonging to 17 meta categories. For each meta category, we select three products for presentation.}
\label{fig:200sku}
\end{figure*}

In this paper, we propose the RPC dataset to support research on approaches to address the potential challenges in real-world ACO scenarios. The characteristics of the dataset can be summarized into six aspects.
\begin{itemize}
\itemsep-0.2em
\item\textbf{Large-scale:}
As shown in Fig.~\ref{fig:plot}, our RPC dataset is the largest dataset so far for retail ACO in terms of product categories (stock keeping units or SKUs) and product images. To collect this dataset, we choose 200 SKUs and purchase on average 4 instances for each SKU, which almost doubles the category size of previous largest dataset. 
In total, we capture 83,739 images including 53,739 single-product exemplar images, and 30,000 checkout images.

\item\textbf{Single-product exemplar images and checkout images:}
In our dataset, we collect two types of images. One type is the exemplar image for every single product (Fig.~\ref{fig:singledesk}) and the other type is the checkout image (Fig.~\ref{fig:checkoutimgs}) taken at the checkout counter. While the exemplar images capture the multi-view appearances of the isolated SKU, the checkout images reflect realistic checkout scenarios where each image covers a variant number of product instances.

\item\textbf{Close to realistic checkout scenario:}
During the construction of this dataset, we try our best to mimic the realistic retail checkout scenarios to collect the checkout images. The products are randomly chosen and combined; they are freely placed on the checkout background with random orientations; occlusions and complex clutter are also common in our dataset. 

\item\textbf{Hierarchical structure:}
The hierarchical structure of product categories is another characteristic of our RPC dataset. The 200 SKUs can be categorized as 17 meta-categories which cover diverse appearances, such as bottle-like, box-like, canister-like, bag-like, as shown in Figure \ref{fig:200sku}. The SKUs under each meta-category tend to be fine-grained. The hierarchical structure can be exploited, for example, as auxiliary supervision information for advanced training or evaluation, similar to~\cite{yolo2017redmon}.

\item\textbf{Different clutter levels:}
In this dataset, we split checkout images into three clutter levels based on the number of SKUs and product instances in each image, as shown in Table~\ref{table:checkoutmode}. Such a clutter level annotation enables an in-depth inspection of the model capacities.

\item \textbf{Weak to strong supervision:} \label{sect:annotation}
As shown in Fig.~\ref{fig:supervision}, the checkout images in our dataset are provided with three different types of annotations, representing the weak to strong supervisions: (1) shopping list, which records the SKU category and count of each product instance in the checkout image. This is the weakest level of annotation and can be easily obtained in practice. (2) Point-level annotation, which provides the central position and the SKU category of each product in the checkout image. (3) Bounding boxes, which provide bounding box and SKU category for each product. This is the most labor-intensive annotation. The introduction of different types of annotations further enriches the research directions that can be derived from this dataset, \eg, research on weakly supervised detection.

\end{itemize}

\begin{table}[t!]
	\caption{Comparisons with the other related datasets.} \label{table:datasetcomparison}
	\centering
	\small
	\setlength{\tabcolsep}{0.3pt}
	\begin{tabular}{|l|c|c|c|c|}
		\hline
		Datasets           & $\sharp$ categories & $\sharp$ images & $\sharp$ objects & $\sharp$ obj/img \\
		\hline 
		SOIL-47~\cite{koubaroulis2002evaluating}           & 47        & 987      &  --        &  --\\
		Supermarket~\cite{rocha2010automatic}       & 15        & 2,633    &  --        &  --\\
		Gorzi-120~\cite{merler2007recognizing}         & 120       & 11,870   &  --        &  --\\
		Grocery Products~\cite{george2014recognizing}  & 80        & 9,030    &  --        &  --\\
		Feribur Groceries~\cite{jund2016freiburg} & 25        & 5,021    &  --        &  --\\
		MVTec D2S~\cite{ulrichmvtec}     & 60        & 21,000   &  72,447   &  3.45\\ 
		\hline
		Our RPC dataset  exemplar & \textbf{200}  & \textbf{53,739} & \textbf{53,739} & \textbf{1}\\
		Our RPC dataset  checkout & \textbf{200}  & \textbf{30,000} & \textbf{367,935} & \textbf{12.26} \\
		\hline
\end{tabular}
\end{table}

\subsubsection{Construction details for the proposed RPC dataset}
In our RPC dataset, we collect 200 retail SKUs. The collected SKUs can be divided into 17 meta-categories, \ie, \texttt{puffed food}, \texttt{dried fruit}, \texttt{dried food}, \texttt{instant drink}, \texttt{instant noodles}, \texttt{dessert}, \texttt{drink}, \texttt{alcohol}, \texttt{milk}, \texttt{canned food}, \texttt{chocolate}, \texttt{gum}, \texttt{candy}, \texttt{seasoner}, \texttt{personal hygiene}, \texttt{tissue}, \texttt{stationery}. Fig.~\ref{fig:200sku} shows some examples for each of these meta-categories. As can be seen, the dataset covers products with diverse appearances and shapes such as bottle-like, box-like, canister-like, bag-like, to name a few. At the same time, the products under same meta-category normally tend to be fine-grained. This constitutes one of the challenges for ACO system.

We collect 53,739 single-product images for isolated SKUs as exemplar images and 30,000 checkout images for ACO system evaluation. Table~\ref{table:datasetcomparison} shows a detailed comparison between our RPC dataset and existing relevant datasets. We introduce the construction details for these two types of images in the following parts.


\paragraph{Construction for single-product exemplar images:}

\begin{figure}[t!]
	\centering
	\subfloat[Examples of bottle-like SKUs.]  { \includegraphics[width=0.45\columnwidth,,height=10.5em]{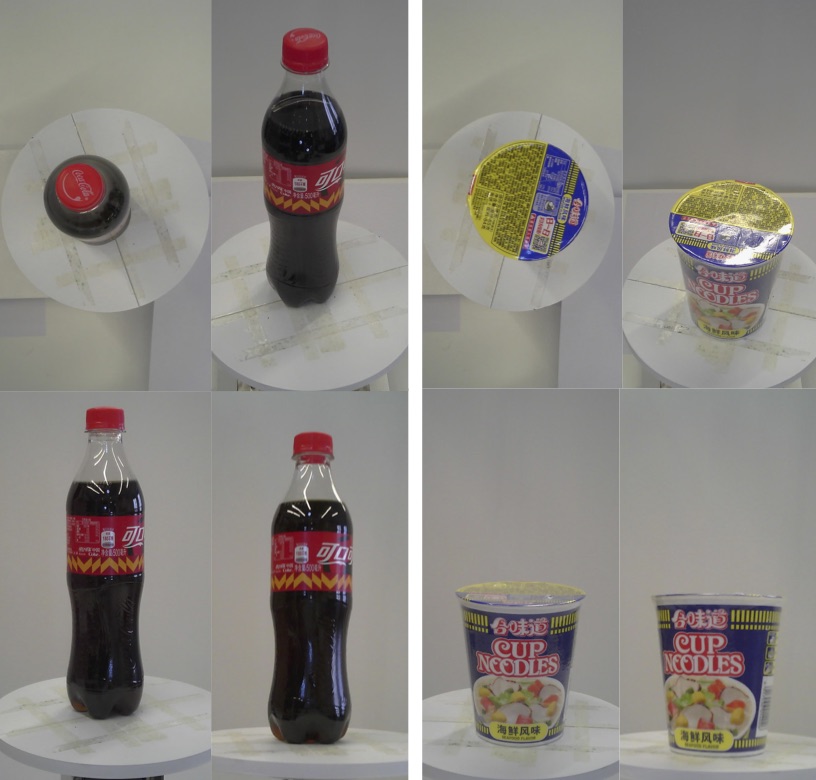} \label{fig:examplesku4view1} } \quad
	\subfloat[Examples of bag-like SKUs.] { \includegraphics[width=0.45\columnwidth,height=10.5em]{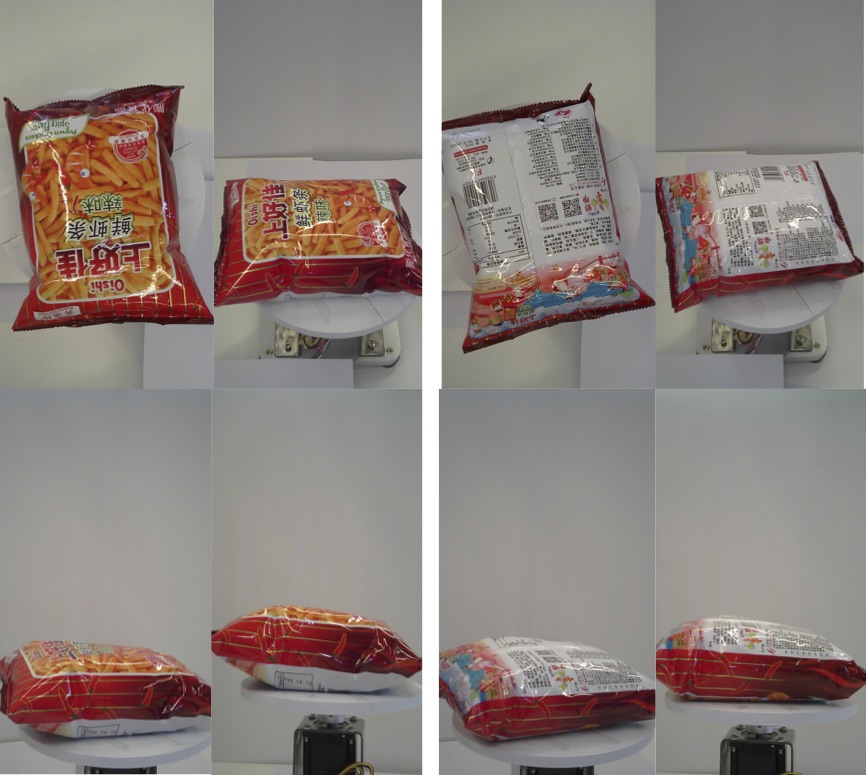} \label{fig:examplesku4view2} }
	\vspace{0.5em}
	\caption{Sampled images of single products. Note that, for bag-like and box-like SKUs, we collect both front and back appearance images.} \label{fig:examplesku4view}
\end{figure}

Fig.~\ref{fig:singledesk} shows the collection environment for capturing the exemplar images. To capture multi-view characteristics for each isolated SKU, four cameras are mounted in different positions to capture images from four different views. One camera covers the top view, one camera covers the horizontal view, and the other two cameras cover the $45^\circ$ and $30^\circ$ views, respectively. The cameras are with automatic focus and capture images with resolution $2592\times 1944$. We randomly choose one instance from each SKU and place it on a turntable which can rotate 360 degrees. Each camera takes a photo for this SKU every 9 degrees. Thus, we have $4\times \left(360/9\right) = 160$ views for each SKU. In addition, for box-like and bag-like SKUs, because their top view is normally different from the bottom view, we repeat the above collecting procedure twice to collect images for both sides. Some example collections from the four cameras are shown in Fig.~\ref{fig:examplesku4view}. In total, we collect 53,739 exemplar images for 200 isolated SKUs.

\begin{figure}[t!]
\centering
	{\includegraphics[width=0.5\columnwidth]{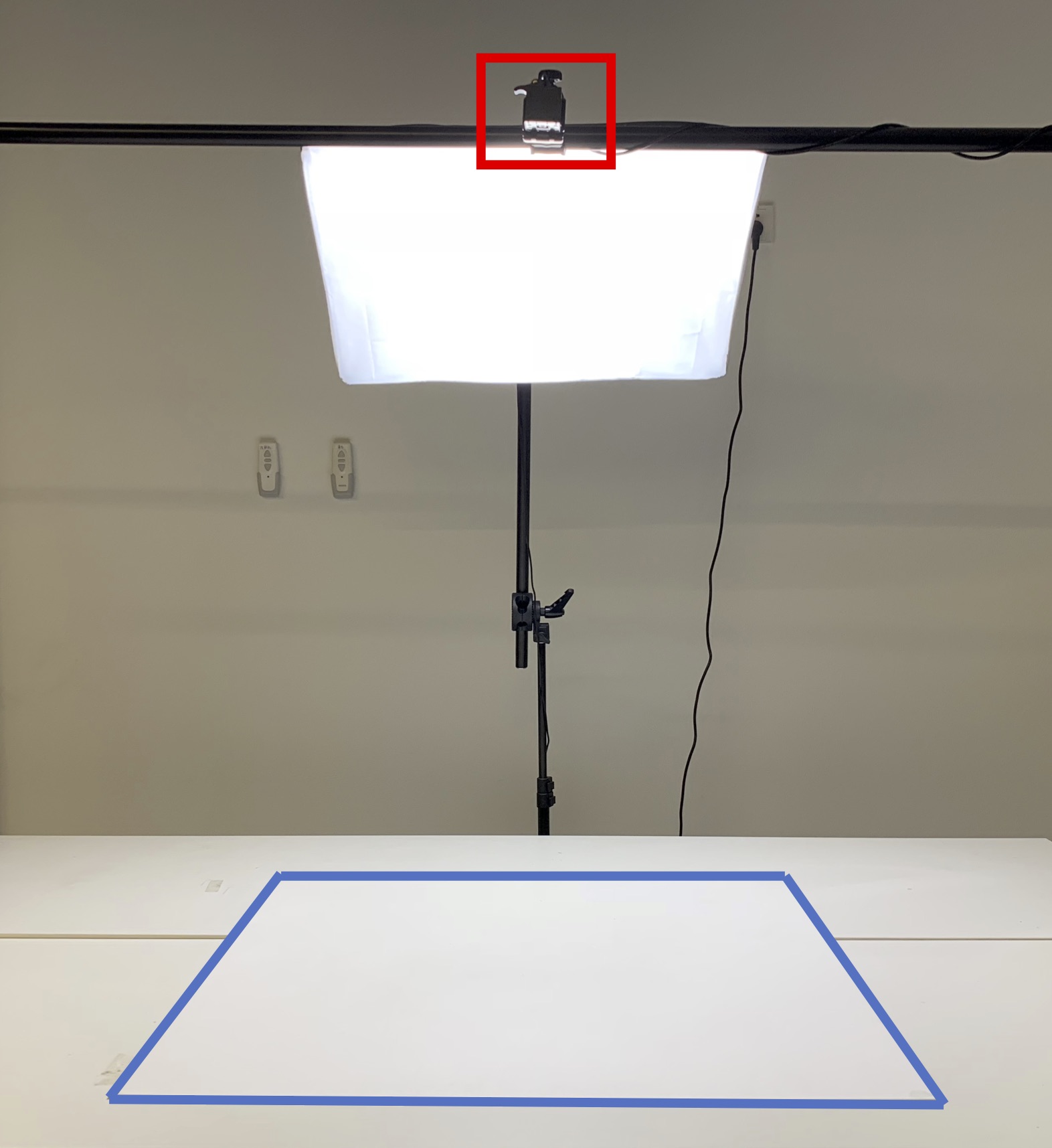}}
\caption{Collection environment of checkout images. Red rectangle marks the camera, and the blue quadrangle is the white board as checkout background.}
\label{fig:checkoutdesk}
\end{figure}

\paragraph{Construction for checkout images:} To capture the checkout images, the products are placed on a $80$cm$\times$$80$cm white board as background with a camera (with resolution $1800\times 1800$) mounted on top. The collection environment is shown in Fig.~\ref{fig:checkoutdesk}. Based on the number of SKUs as well as the number of product instances for each SKU, we collect checkout images with three clutter levels: \ie, \texttt{easy}, \texttt{medium} and \texttt{hard}. Table~\ref{table:checkoutmode} shows the details of the three splits. Generally, the more products presented on the board, the more challenging thing is to accurately recognize the whole set of products due to problems such as occlusion, varying orientations and complex clutter and density patterns, as shown in Fig.~\ref{fig:checkoutimgs}. To capture a checkout image, we mimic the realistic checkout scenario. Concretely, we select a random set of SKUs and a random number of product instances for each of the selected SKUs guided by specific clutter level (Table~\ref{table:checkoutmode}), and freely place these products onto the board. To capture comprehensive product combinations, for each clutter level, we repeat this process 10,000 times to collect 10,000 images. Thus in total, we collect 30,000 checkout images.

%

\begin{table}[t!]
	\caption{Three clutter levels of checkout images.} \label{table:checkoutmode}
	\centering
	\small
	\begin{tabular}{|l||c|c|}
		\hline
		{Clutter levels}  & {$\sharp$ categories} & {$\sharp$ instances} \\  
		\hline
		Easy mode & $3\sim 5$ &  $3\sim 10$ \\
		Medium mode & $5\sim 8$ & $10\sim 15$ \\
		Hard mode & $8\sim 10$ & $15\sim 20$ \\
		\hline
	\end{tabular}
\end{table}

\subsection{Evaluation protocol}\label{sec:protocol}

In this section, we propose several evaluation metrics for the proposed ACO task. First, we introduce some notations for clear presentations.

Given $N$ checkout images from $K$ SKU categories, let $P_{i,k}$ denote the predicted count of the $k$-th category in the $i$-th image and $GT_{i,k}$ represent the ground truth instance number of the $k$-th category in the $i$-th image. $CD_{i,k}$ is defined as the $\ell_1$ distance between the prediction $P_{i,k}$ and the ground truth $GT_{i,k}$, which reflects the counting error for a specific SKU category in an image,
\begin{equation}
CD_{i,k} = |P_{i,k}- GT_{i,k}|\,.
\end{equation}
Then, $CD_{i}$ is defined as the prediction error over all $K$ SKU categories for the $i$-th image,

\begin{equation}
CD_{i} = \sum_{k=1}^{K}{CD_{i,k}},    
\end{equation}
where $CD_i=0$ indicates the complete product list of a checkout image is correctly predicted.

\paragraph{Checkout Accuracy (cAcc):}

Checkout accuracy is the most important metric for the ACO task. It measures the pass rate of a ACO system and thus reflects the practicality of the system. Under this metric, an image passes checkout if and only if the complete product list is accurately predicted, \ie, $CD_i=0$.
Mathematically, this metric can be written as,

\begin{equation}
{\rm cAcc} = \frac{\sum_{i=1}^{N}\delta\left(\sum_{k=1}^{K}CD_{i,k},\quad 0\right)}{N} \,,
\end{equation}
where $\delta(\cdot)$ returns 1 if and only if $\sum_{k=1}^{K}CD_{i,k}=0$; otherwise, it returns 0. The range of the cAcc score is $\left[0, 1\right]$.

\paragraph{Average Counting Distance (ACD):}

Different from cACC that only cares whether the complete product list is correctly predicted or not, Average Counting Distance (ACD) measures the average number
of counting errors for each image.

\begin{equation}
{\rm ACD} = \frac{1}{N}\sum_{i=1}^{N}\sum_{k=1}^{K}CD_{i,k} \,.
\end{equation}

\paragraph{Mean Category Counting Distance (mCCD):}

Mean Category Counting Distance (mCCD) is proposed to measure the average ratio of counting errors for each SKU category,

\begin{equation}
{\rm mCCD} = \frac{1}{K}\sum_{k=1}^{K} \frac{\sum_{i=1}^{N}CD_{i,k}}{\sum_{i=1}^{N}GT_{i,k}} \,.
\end{equation}

\paragraph{Mean Category Intersection of Union (mCIoU):}

Mean Category Intersection of Union (mCIoU) is another metric proposed to measure the compatibility between the predicted shopping list and ground truth. It is motivated by standard IoU. The range of the mCIoU score is $\left[0, 1\right]$.

\begin{equation}
\label{eq:mciou}
{\rm mCIoU} = \frac{1}{K}\sum_{k=1}^{K} \frac{\sum_{i=1}^{N}\min \left(GT_{i,k}, P_{i,k}\right)}{\sum_{i=1}^{N} \max \left(GT_{i,k}, P_{i,k}\right)} \,.
\end{equation}

\section{Benchmarking the proposed RPC dataset}\label{sec:method}
\subsection{Our baseline solutions to the ACO problem}
The proposed ACO problem is an open problem and has many potential solutions. To benchmark the proposed ACO dataset, in this section, we consider four baseline approaches which formulate the ACO problem as a cross-domain detection problem. We restricted ourselves to only use the annotation from the single-product exemplar images and the most straightforward way is to directly train detectors on these exemplar images. We use this strategy as our first baseline (denoted as \textbf{Single}) and we adopt Feature Pyramid Network (FPN) \cite{lin2017feature} as the detector. 

The checkout image contains multiple objects while the exemplar image only has one. To reduce this gap, we propose to \emph{copy and paste} the segmented isolated products to create synthesized checkout images. Then we can train the detector on the synthesized image (100,000 images are synthesized) and we use this approach as our second baseline (denoted as \textbf{Syn}). To segment the product instance, we adopt a salience-based object segmentation approach~\cite{ping2016detectingTIP} with conditional random fields~\cite{krahenbuhl2011efficient} for mask refinement. For more details about the creation of the synthesized checkout images, please refer to the supplementary materials. 

\begin{figure}[t!]
\centering
	{\includegraphics[width=0.7\columnwidth]{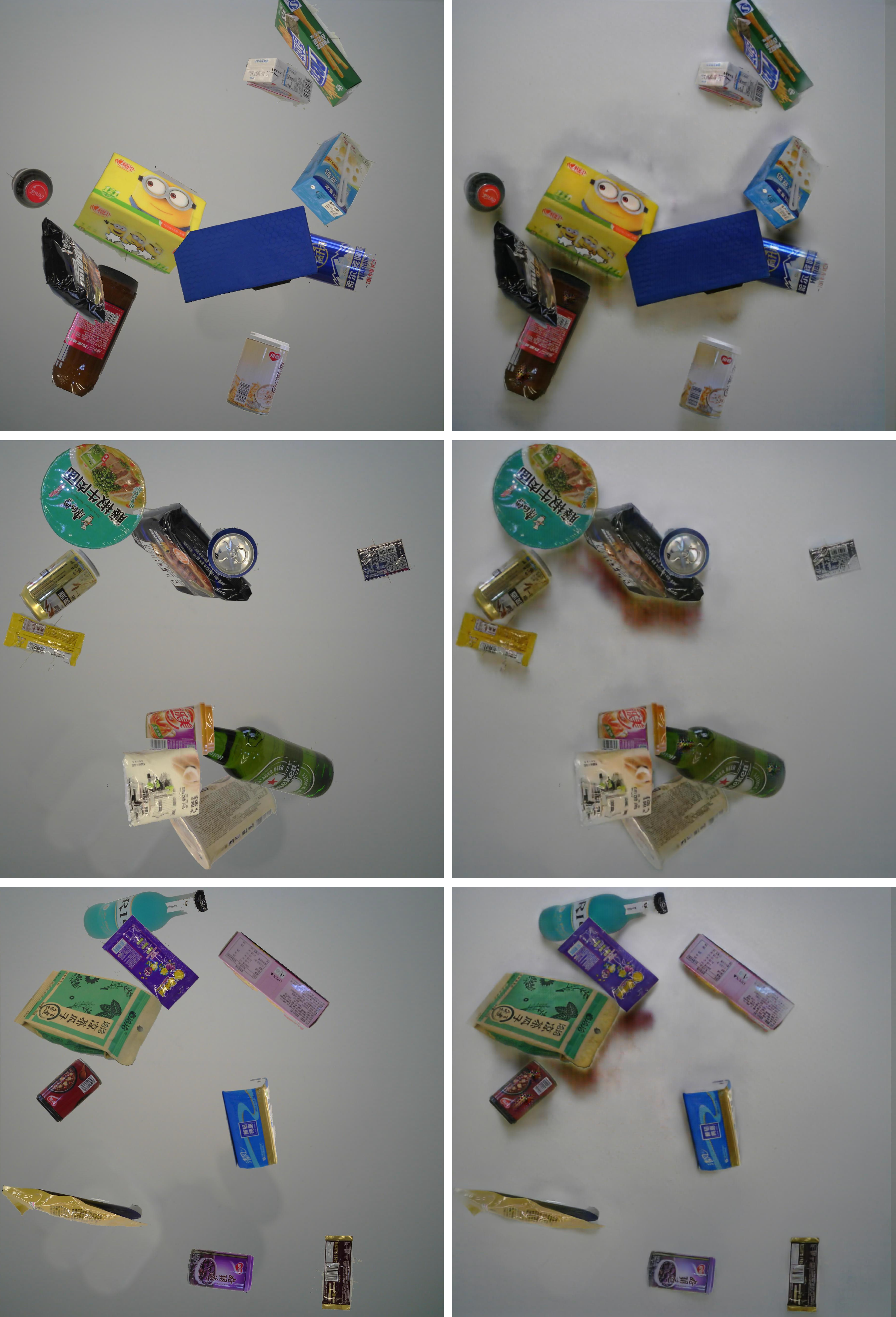}}
\caption{Examples of the synthesized checkout images (on the left) and the corresponding images rendered (on the right) by Cycle-GAN~\cite{zhu2017unpaired}.}
\label{fig:synsandtrans}
\end{figure}

\begin{figure*}[t!]
\centering
\vspace{-2em}
	{\includegraphics[width=0.78\textwidth]{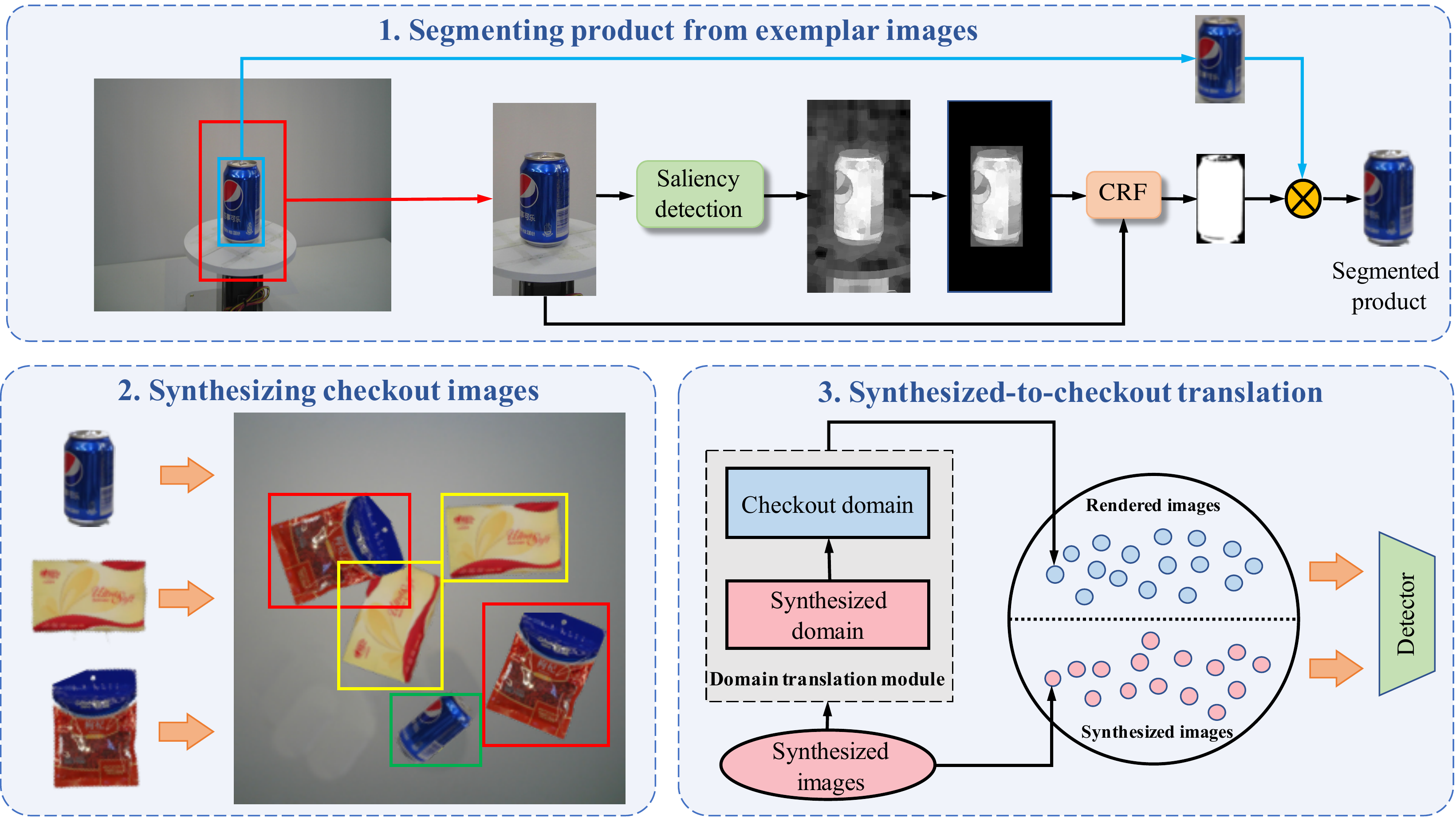}}
\caption{Pipeline of our proposed method for the ACO task.}
\label{fig:proposedmethod}
\end{figure*}

After the synthesis step, domain gap still exists between the synthesized images and checkout images. It is easy to tell the difference between the images from these two domains by observing lighting conditions or shadow patterns. In order to render the synthesized images more naturally similar to checkout images, we employ Cycle-GAN~\cite{zhu2017unpaired} to translate these images into the checkout image domain. Fig.~\ref{fig:synsandtrans} shows the synthesized checkout images and the corresponding translated images by Cycle-GAN. It can be seen that the rendered images are more realistic. Then, we train the detector on the rendered images (rendering on 100,000 synthesized images), which is regarded as the third baseline (denoted as \textbf{Render}). In addition, we also train detectors with both the rendered images and the synthesized ones. We use this as our final baseline and denote it as \textbf{Syn+Render}. The pipeline of this method is shown in Fig.~\ref{fig:proposedmethod}.



\subsection{Experimental evaluation}



\paragraph{Evaluation settings and implementation details}
Both the evaluation metrics proposed in Sec.~\ref{sec:protocol}, and the standard object detection metrics, \ie, mAP50 and mmAP, are adopted to evaluate the above methods.


For product detector training, an input image is resized such that its shorter side has 800 pixels. Synchronized SGD is used to train the model on 2 GPUs. Each mini-batch consists of 2 images on each GPU and we set the number of RoIs to be 512 for each image. We set the weight decay to be 0.0001 and momentum is set to be 0.9. The initial learning rate is 0.02 for the first 60k iterations, which decays by a factor of 10 for the next 20k iterations. Note that for each clutter level, we use 2,000 checkout images as validation set and the remaining 8,000 images are used as the test set.

\begin{table*}[t!]
\caption{Experimental results of the ACO task on our RPC dataset.} \label{table:results}
\centering
\footnotesize
\begin{tabular}{|c|c||r|r|r|r|r|r|}
\hline
\textit{Clutter mode} & \textit{Methods} & \textit{cAcc} ($\uparrow$) & \textit{ACD} ($\downarrow$) & \textit{mCCD} ($\downarrow$) & \textit{mCIoU} ($\uparrow$) & \textit{mAP50} ($\uparrow$) & \textit{mmAP} ($\uparrow$) \\
\hline
    \hline
\multirow{4}{*}{\texttt{Easy}} & Single & 0.02\% & 7.83 & 1.09 & 4.36\% & 3.65\% & 2.04\% \\
	& Syn & 18.49\%  &  2.58  &  0.37  &  69.33\%  &  81.51\%  &  56.39\% \\
        & Render & 63.19\%  &  0.72  &  0.11  &  90.64\%  &  96.21\%  &  77.65\% \\
        & Syn+Render & \textbf{73.17\%}  &  \textbf{0.49}  &  \textbf{0.07}  &  \textbf{93.66\%}  &  \textbf{97.34\%}  &  \textbf{79.01\%} \\
\hline
\multirow{4}{*}{\texttt{Medium}} & Single & 0.00\% & 19.77 & 1.67 & 3.96\% & 2.06\% & 1.11\% \\
	& Syn & 6.54\%  &  4.33  &  0.37  &  68.61\%  &  79.72\%  &  51.75\% \\
        & Render & 43.02\%  &  1.24  &  0.11  &  90.64\%  &  95.83\%  &  72.53\% \\
        & Syn+Render & \textbf{54.69\%}  &  \textbf{0.90}  &  \textbf{0.08}  &  \textbf{92.95\%}  &  \textbf{96.56\%}  &  \textbf{73.24\%} \\
\hline
\multirow{4}{*}{\texttt{Hard}} & Single & 0.00\% & 22.61 & 1.33 & 2.06\% & 0.97\% & 0.55\% \\
	& Syn & 2.91\%  &  5.94  &  0.34  &  70.25\%  &  80.98\%  &  53.11\% \\
        & Render & 31.01\%  &  1.77  &  0.10  &  90.41\%  &  95.18\%  &  71.56\% \\
        & Syn+Render & \textbf{42.48\%}  &  \textbf{1.28}  &  \textbf{0.07}  &  \textbf{93.06\%}  &  \textbf{96.45\%}  &  \textbf{72.72\%} \\
\hline
\hline
\multirow{4}{*}{Averaged} & Single & 0.01\% & 12.84 & 1.06 & 2.14\% & 1.83\% & 1.01\% \\
        & Syn & 9.27\%  &  4.27  &  0.35  &  69.65\%  &  80.66\%  &  53.08\% \\
        & Render & 45.60\%  &  1.25  &  0.10  &  90.58\%  &  95.50\%  &  72.76\% \\
        & Syn+Render & \textbf{56.68\%}  &  \textbf{0.89}  &  \textbf{0.07}  &  \textbf{93.19\%}  &  \textbf{96.57\%}  &  \textbf{73.83\%} \\
\hline
\end{tabular}
\end{table*}

\paragraph{Experimental results}

In this section, we report the experimental results by clutter modes. That is we report the results for \texttt{easy mode, medium mode, hard mode} separately. Table~\ref{table:results} shows the quantitative results.

As can be seen, due to significant domain shift between single-product exemplar images and checkout images, the baseline method that directly trains the product detector using single-product images almost fails the task, even for the \texttt{easy mode}. Improvement is observed when synthesized images are used for product detector training, especially for \texttt{easy}, where the most important metric cAcc is improved by about 18\%. The improvement, however, is not that significant for \texttt{medium} and \texttt{hard} modes. Further boost is achieved by training detectors on rendered images. Since the domain gap is further mitigated via domain translation, the detectors trained on rendered images generalize better to the checkout images.
The cAcc score is improved by 44.70\%, 36.48\%, 28.10\% for \texttt{easy}, \texttt{medium} and \texttt{hard}, respectively. This proves the effectiveness of the domain translation component in our approach.

The most significant performance boost is obtained when the rendered images are combined with synthesized images for detector training. As seen, comparing to methods based on synthesized (rendered) images only, the cAcc is improved by about 55\% (10\%), 48\% (12\%), and 40\% (11\%) for \texttt{easy}, \texttt{medium} and \texttt{hard}, respectively. The observations reveal that synthesis variability is beneficial for the generalization capacity of the model trained on synthetic data. This observation is consistent with that in~\cite{tobin2017domain}. 

However, also as shown in Table~\ref{table:results}, even the best-performed method achieves unsatisfactory performance on the \texttt{medium} and \texttt{hard} modes. This shows the task is challenging and leaves substantial room for improvement. Also, another point worth mentioning is that with standard object detection metrics, the performance of the best performed approach is promising, \eg, 72.72\% mmAP on \texttt{hard mode}. However, from the perspective of practicality, we define a much more strict evaluation metric which requires models to be able to accurately predict the whole product list.

Additionally, we go through the whole checkout test images, and find there are mainly four types of failure cases: (1) missed detection; (2) failure cases caused by dense placement; (3) failure cases caused by fine-grained differences; (4) false positives. Examples of failure and successful cases can be found in the supplementary materials.

\section{Possible research directions on our dataset}

Although we tackle the ACO task with a cross-domain detection strategy in our benchmark, there are many other possible solutions. Moreover, other possible research directions can be derived from the proposed dataset. We name a few of them in this section:
\begin{itemize}
\itemsep-0.3em 
    \item Online learning for the ACO problem. One challenge of the real-world ACO problem is that the new product will be continuously added to the product list. Thus it is desirable to find a way to quickly update the system without retraining the model from scratch. This problem falls into the field of online learning. However, the cross-domain and fine-grained nature of the ACO problem will add extra difficulty to the problem and require the development of new approaches. 
    \item Another potential solution to the ACO task is to directly predict the product list from the checkout image without recursing to the accurate product detection and thus relieves the burden of training a detector. This solution essentially models the ACO problem as an object counting problem \cite{ocounting}. However, it is a new type of object counting problem as it involves objects from multiple categories, and each object has a limited number of training samples. 
    \item Using mixed supervision from the checkout images. Our dataset provides different levels of supervision for the checkout images. How to leverage those annotations for better solving the ACO task is still an open problem and needs more in-depth research. 
    \item As a complementary dataset for other computer vision tasks. Although our dataset is designed for the ACO task, it can also act as a dataset for research areas such as object retrieval, few-shot/weakly-supervised/fully-supervised object detection, since our annotations also include the ground truth location/bounding-box of products in the checkout images.
\end{itemize}
\vspace{-0.4cm}
\section{Conclusion}
In this paper, we proposed a new dataset for the automatic checkout (ACO) task. This dataset contains 200 product categories and 83,739 images. It includes single-product images taken in controlled environment and multi-product checkout images taken at the checkout counter. Various annotations are provided for both single-product images and checkout images. With this dataset, we clearly define the ACO problem and benchmark the dataset with four detection-based baselines. We show that there is still substantial room to improve the ACO performance on this dataset and this dataset can support various potential research directions.

\onecolumn
\appendix
\section*{SUPPLEMENTARY MATERIALS}
In the supplementary materials, we present additional information about the proposed Retail Product Checkout (RPC) dataset, including:

\begin{enumerate}
\item[A.] Example images of all 200 products from the dataset;
\item[B.] Detailed statistic information about the dataset;
\item[C.] Additional examples of checkout images from the dataset;
\item[D.] Details of checkout image synthesis and checkout image generation via domain translation;
\item[E.] Failure and successful cases of the ACO task on the dataset;
\item[F.] The ACO performance \vs the number of synthesized and rendered images via domain translation.
\item[G.] Detailed statistical ACO results on single meta-category and single product category
\end{enumerate}

\clearpage
\section{Example images of all 200 products from the dataset}
\begin{figure*}[h!]
	\centering
	\includegraphics[width=0.92\textwidth]{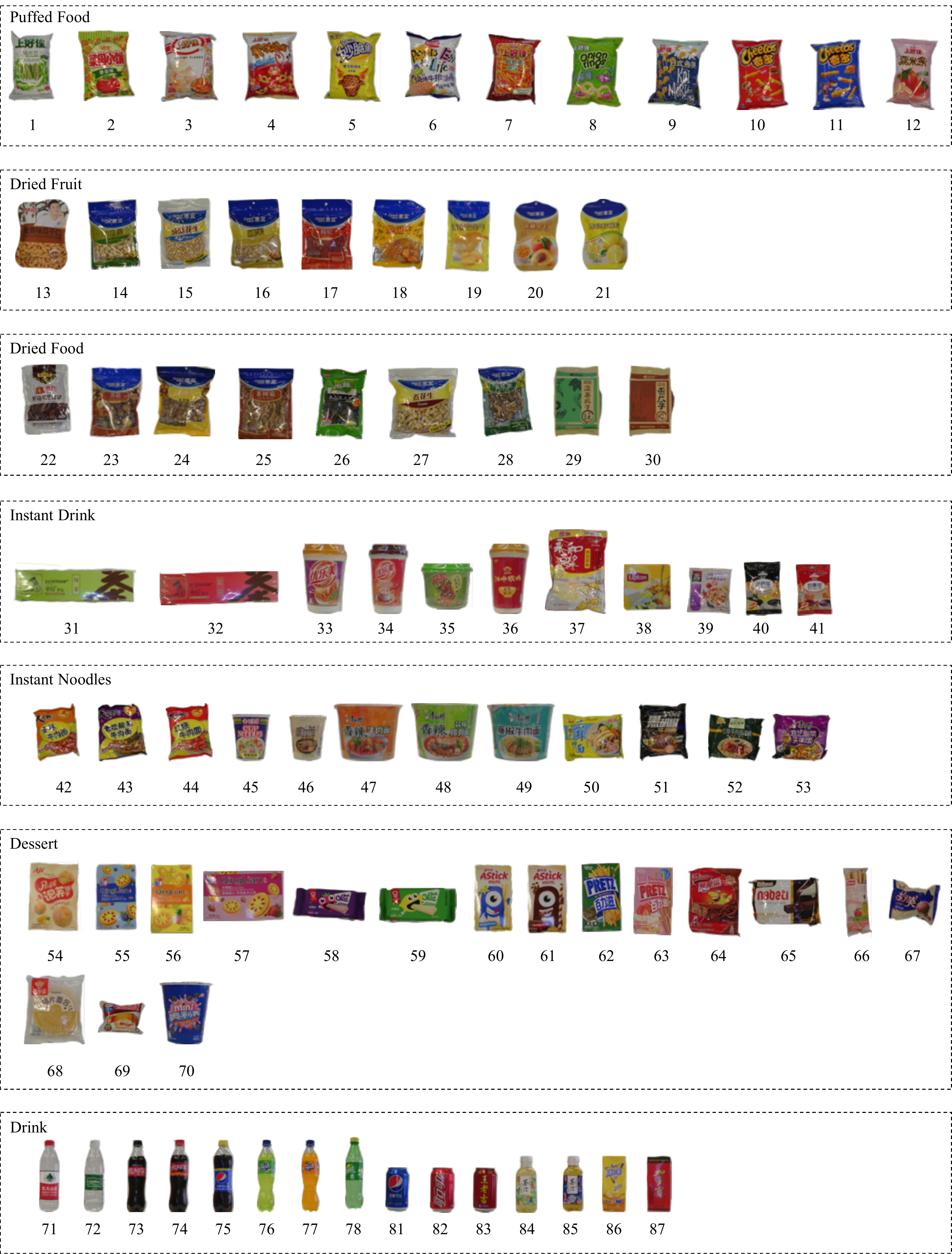}\label{fig:200sku_1}
\end{figure*}

\begin{figure*}[p!]
	\centering
	\includegraphics[width=0.92\textwidth]{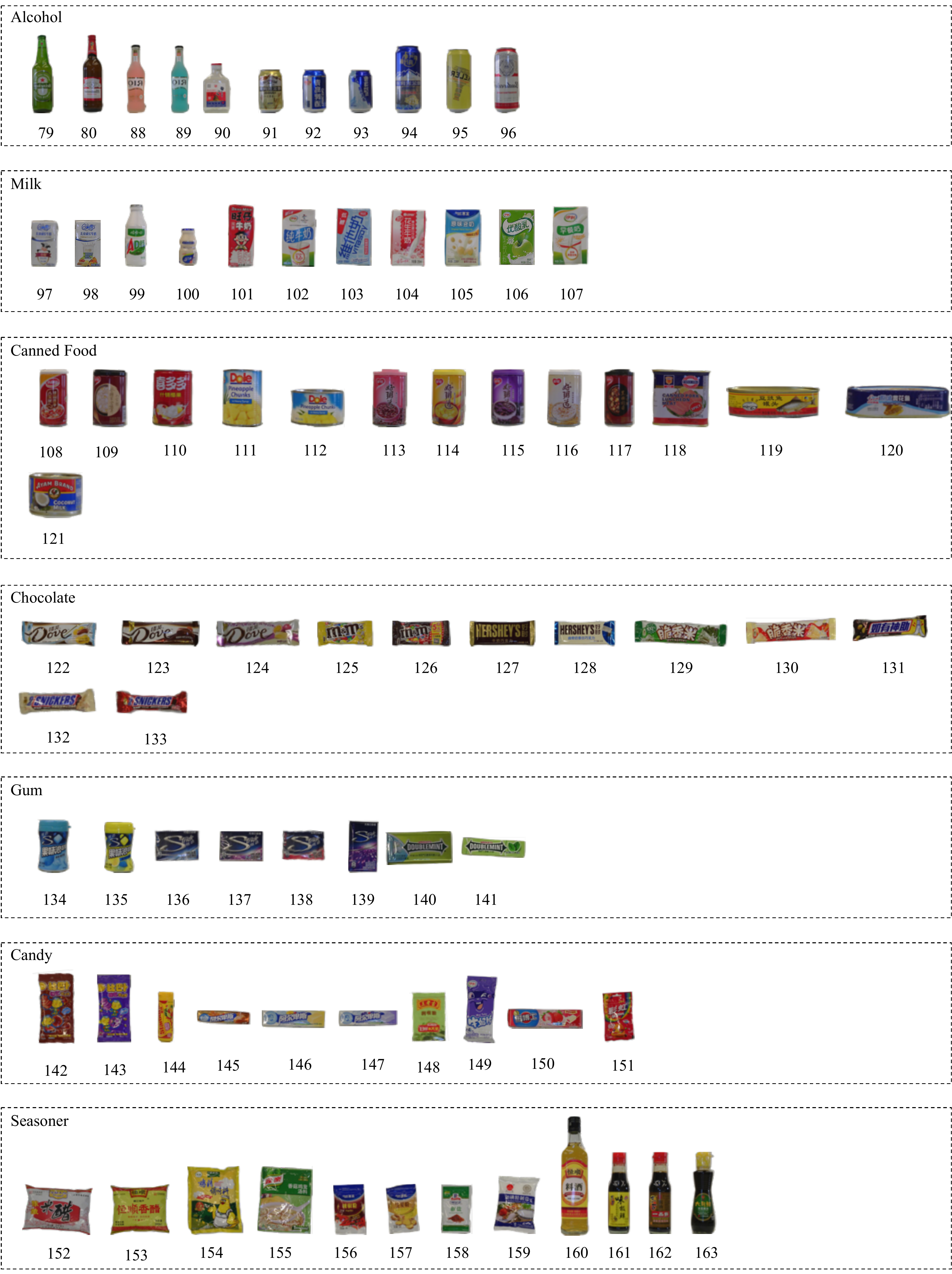}\label{fig:200sku_2}
\end{figure*}

\begin{figure*}[p!]
	\centering
	\includegraphics[width=0.92\textwidth]{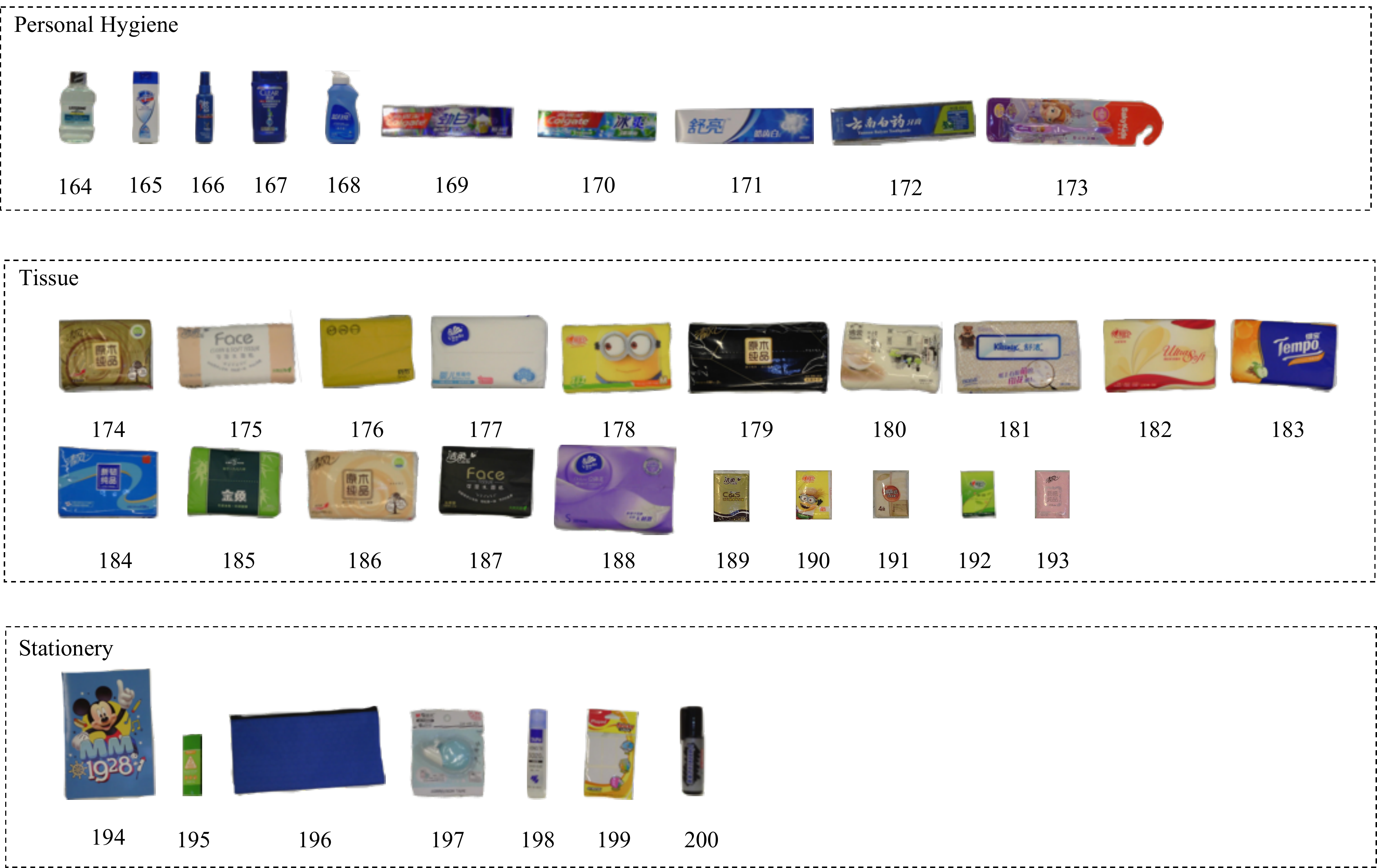}\label{fig:200sku_3}
	\vspace{0.5em}
	\caption{Example images of all 200 products from the RPC dataset. The products are organized by meta-category. The number under each example is the corresponding product ID.}
\end{figure*}

\clearpage
\section{Detailed statistic information about the dataset} 
\begin{figure*}[h!]
	\centering
	\includegraphics[width=0.7\textwidth]{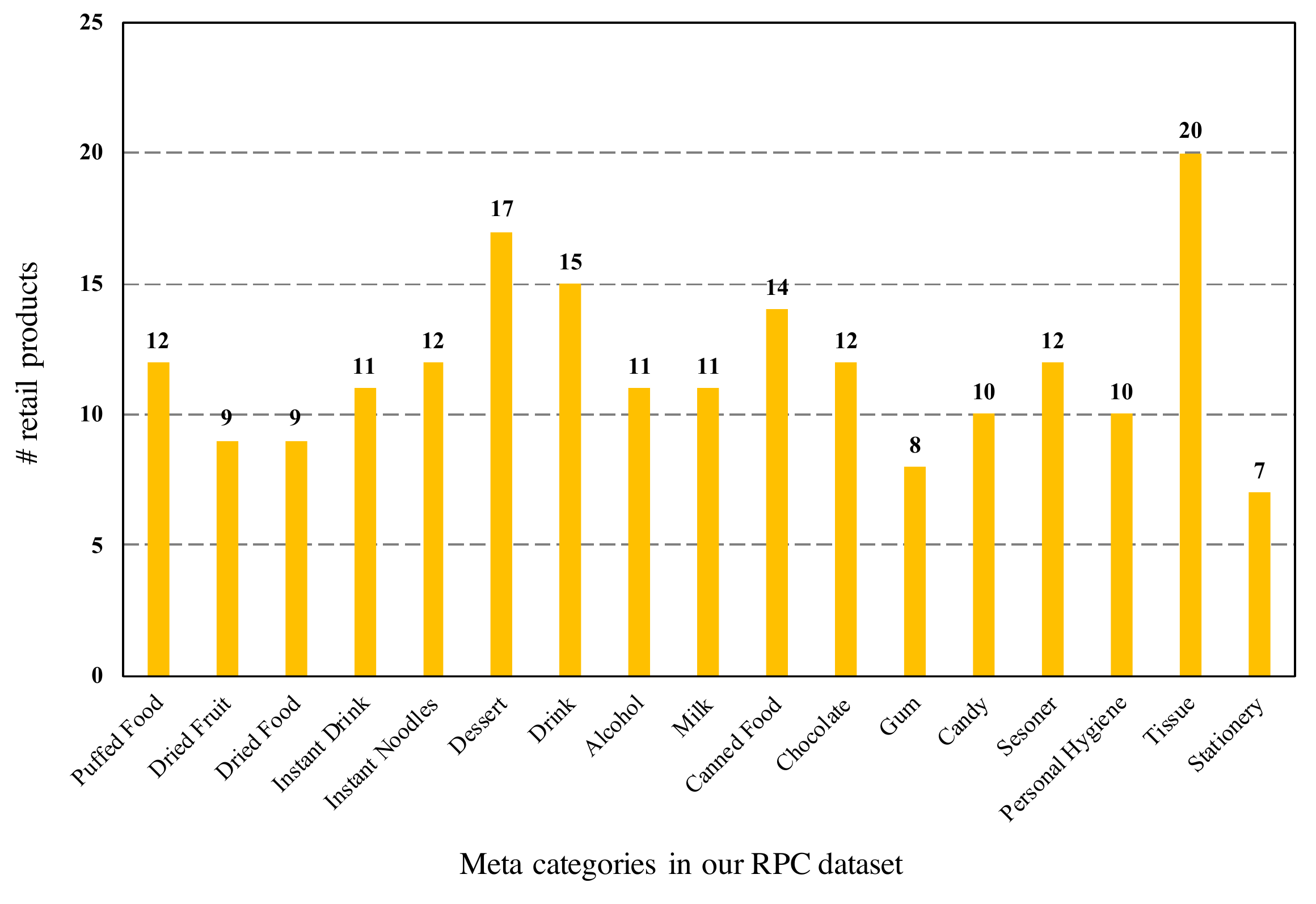} 
	\caption{The number of product categories in each meta-category.}
	\label{fig:numsku_permetaclass}
\end{figure*}

\begin{figure*}[h]
	\centering
	\includegraphics[width=0.25\textwidth]{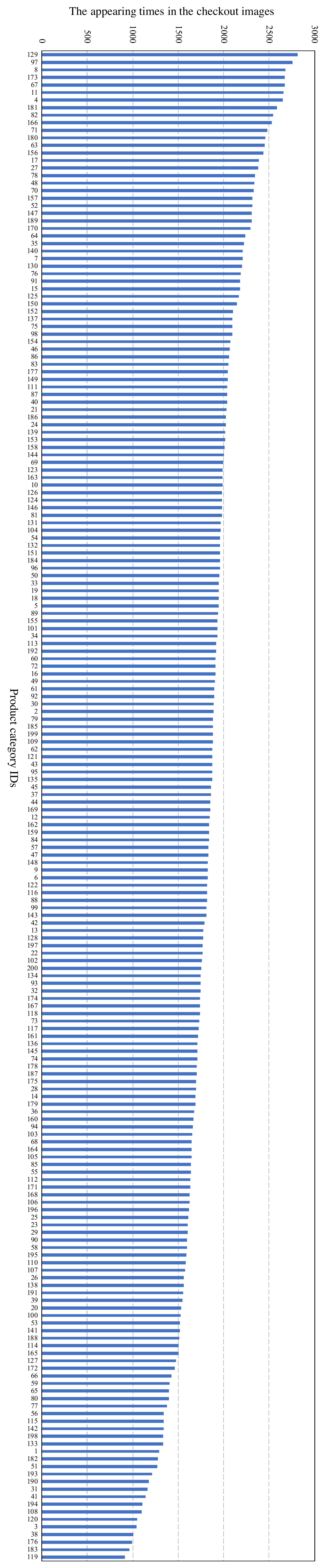}
	\caption{The number of times each product category appears in the checkout images of the proposed RPC dataset.}
	\label{fig:numberpersku}
\end{figure*}

\clearpage
\section{Additional examples of checkout images from the dataset}

In this section, we show additional checkout images of three different clutter modes (\ie, \texttt{Easy}, \texttt{Medium} and \texttt{Hard}) from our RPC dataset.

\begin{figure*}[h!]
 \centering

 \subfloat[\texttt{Easy} clutter mode checkout images.]  { \includegraphics[width=0.95\textwidth]{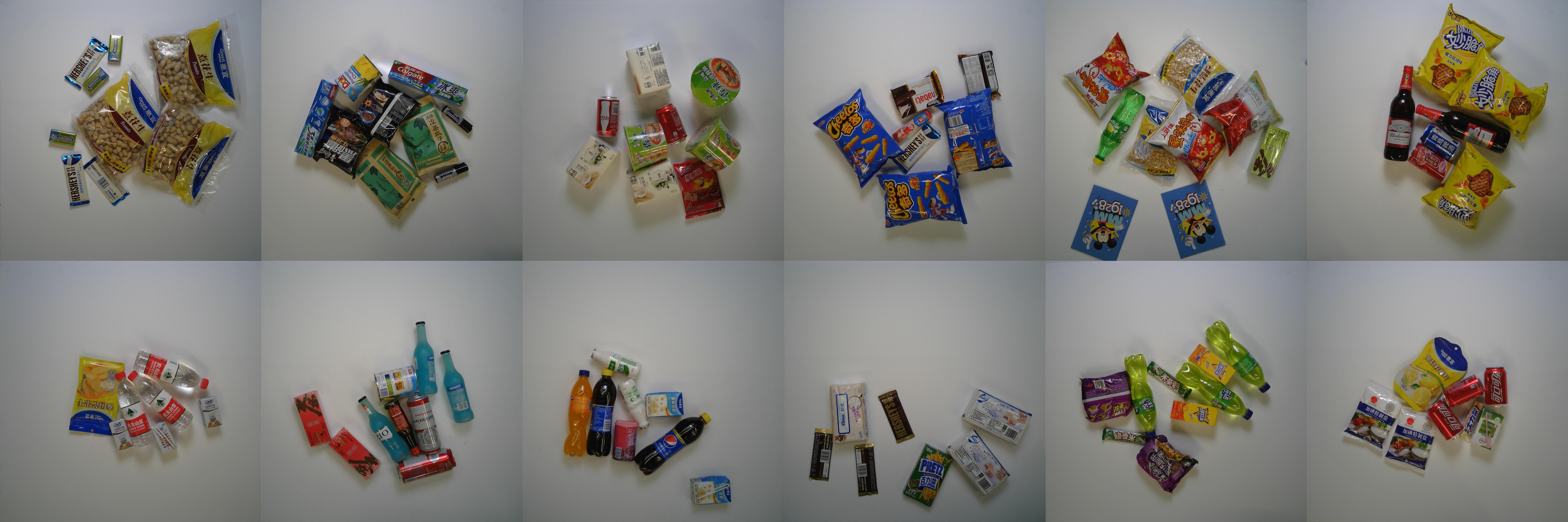} \label{fig:appendix_easy} } \\
 
 \subfloat[\texttt{Medium} clutte mode checkout images.]  { \includegraphics[width=0.95\textwidth]{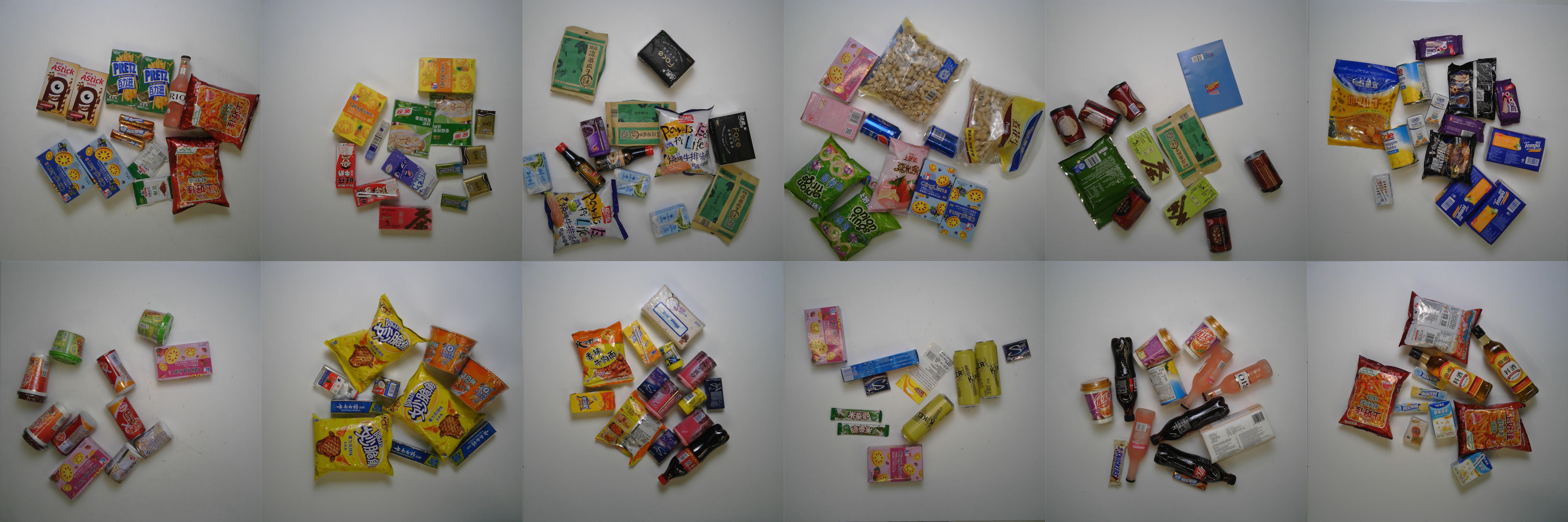} \label{fig:appendix_medium} } \\
 
 \subfloat[\texttt{Hard} clutter mode checkout images.]  { \includegraphics[width=0.95\textwidth]{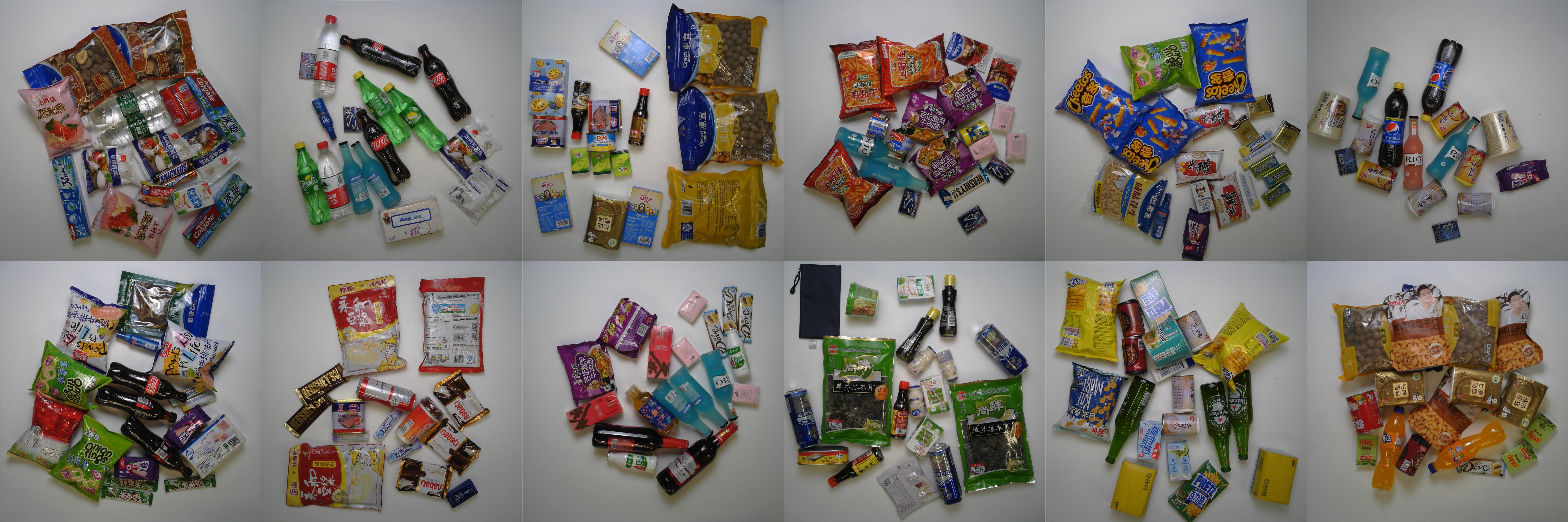} \label{fig:appendix_hard} } \\
 
 \vspace{0.5em}
 \caption{Checkout images of three different clutter modes.} \label{fig:appendix_checkout}
\end{figure*}

\clearpage
\section{Details of checkout images synthesis and checkout image generation via domain translation}

In this section, we describe the detailed process of synthesizing and rendering checkout images.

\subsection{Segmenting SKU from exemplar images}

As shown in Fig.~10 in the paper, to synthesize the checkout images with multiple SKUs, the first step is to segment the isolated SKUs from the exemplar single-product images. We adopt a salience based segmentation approach to realize this. To differentiate the product from the background, the salience method needs to see some contextual background information. Thus we expand the size of the annotated product bounding box by two times (\eg, expanding the blue annotated bounding box for \textit{Pepsi} to the red bounding box in Fig.~1 in the paper) and crop the RoI based on that expanded bounding box. Then, for the cropped image patch, we utilize the method proposed in~\cite{ping2016detectingTIP} to obtain the saliency map as the product mask. After that, we set the value of background pixels which are outside the annotated product bounding box (\ie, the blue one) to be zero. Finally, conditional random field (CRF)~\cite{krahenbuhl2011efficient} is applied to further refine the processed saliency map, and the refined saliency map is used as the mask to segment the SKU.

\subsection{Synthesizing checkout images via a \emph{copy and paste} strategy}

After obtaining the segmented SKUs, we synthesize the checkout images following the idea of \emph{copy and paste}. 
Specifically, segmented SKUs are randomly selected and freely placed on a prepared background image. 
We set the occlusion rate of each SKU less than 50\%. To mimic different clutter modes in checkout images, the SKU number as well as the instance number of each SKU are determined under the same principle shown in Table~2 in the paper. After this step, the synthesized images are similar to the checkout images in terms of products placement.

\subsection{Translating synthesized images to checkout images via domain translation}

\begin{figure*}[h]
	\centering
	\includegraphics[width=0.95\textwidth]{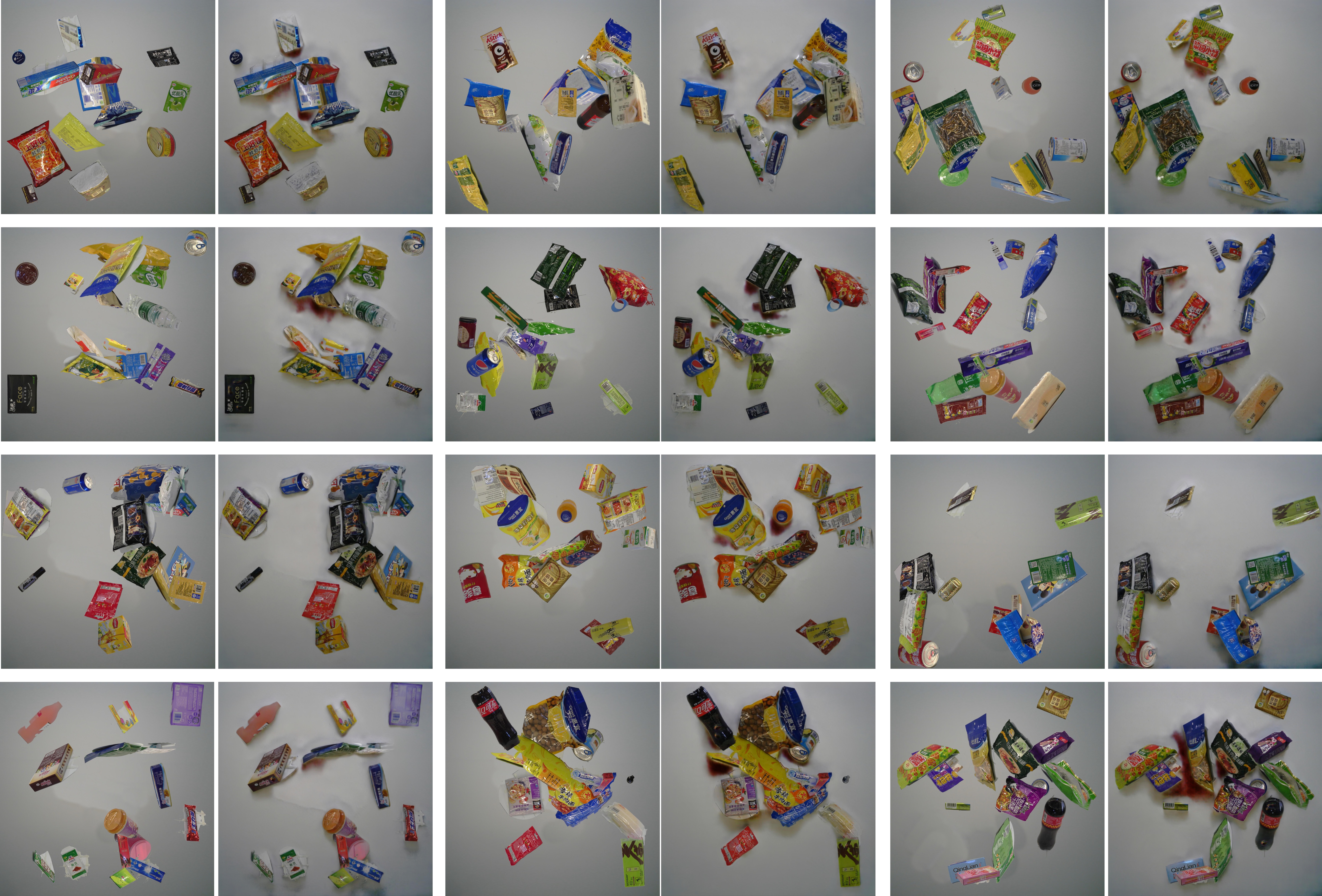}
	\caption{Additional synthesized checkout images (left) and the corresponding images rendered by Cycle-GAN (right).}
	\label{fig:beforeafterDT}
\end{figure*}

After the synthesis step, domain gap still exists between the synthesized images and checkout images, such as the lighting conditions, the shadow patterns, to name a few. In order to render the synthesized images more naturally similar to checkout images, we employ Cycle-GAN~\cite{zhu2017unpaired} to translate these images to the checkout image domain. Fig.~\ref{fig:beforeafterDT} shows more synthesized checkout images and the corresponding translated images by Cycle-GAN. Based on these rendered images, we can train SKU detectors and apply these detectors on real checkout images to predict the complete product list.

More concretely, to render the synthesized images, we follow the experiment settings in~\cite{zhu2017unpaired}. Firstly, we resize the original synthesized images as 0.44x. Then, we use training patches with size $256\times 256$ to train a Cycle-GAN. The objective loss function of the discriminator is defined as a least-square loss~\cite{least2017mao}. For optimization, we use Adam~\cite{adam} optimizer with initial learning rate to be $2\times 10^{-4}$ and the momentum term $\beta_1$ to be 0.5. The total number of training epochs is 200.

\clearpage
\section{Failure and successful cases of the ACO task on the dataset}

\begin{figure*}[h]
 \centering

 \subfloat[\textbf{Missed detection.} The yellow bounding boxes represent the missed detection, and the red bounding boxes are false positives.]  { \includegraphics[width=0.92\textwidth]{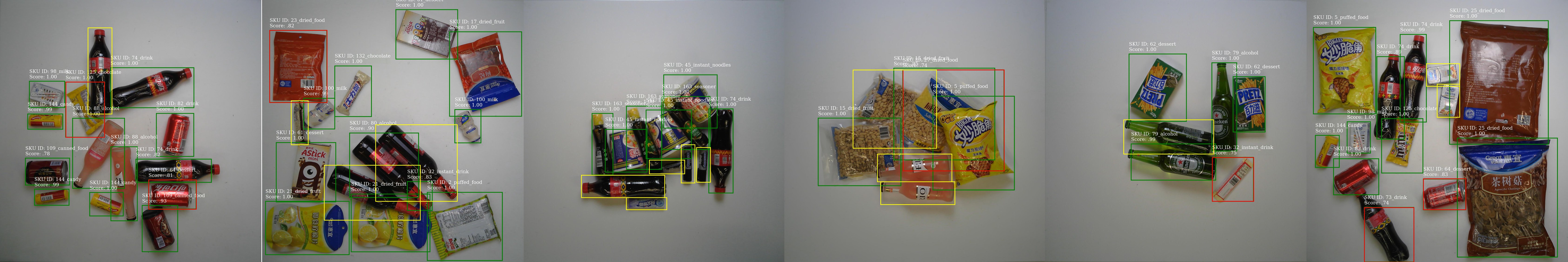} \label{fig:detmiss} } \\
 
 \subfloat[\textbf{Failure cases caused by dense placement.} The yellow bounding boxes represent the missed detection, and the red are false positives.]  { \includegraphics[width=0.92\textwidth]{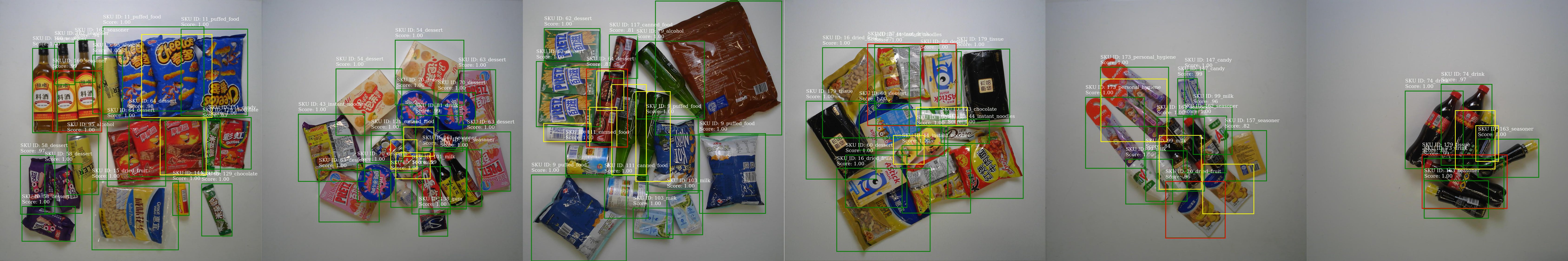} \label{fig:detcrowd} } \\
 
 \subfloat[\textbf{Failure cases caused by fine-grained differences.} The yellow bounding boxes represent the missed detection, while the red bounding boxes are the wrong predictions caused by fine-grained differences. We also present the predictions with the corresponding ground truth labels.]  { \includegraphics[width=0.92\textwidth]{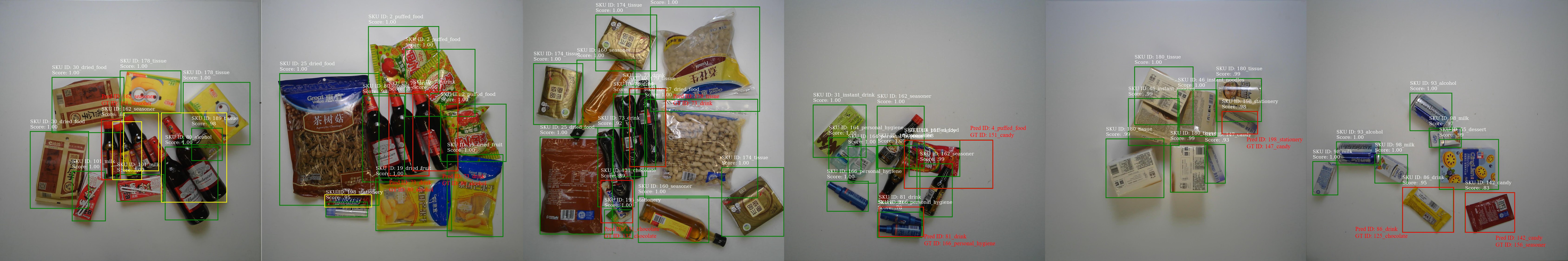} \label{fig:detfg} } \\
 
 \subfloat[\textbf{False positives.} The yellow bounding boxes represent the missed detection, and the red bounding boxes are false positives.]  { \includegraphics[width=0.92\textwidth]{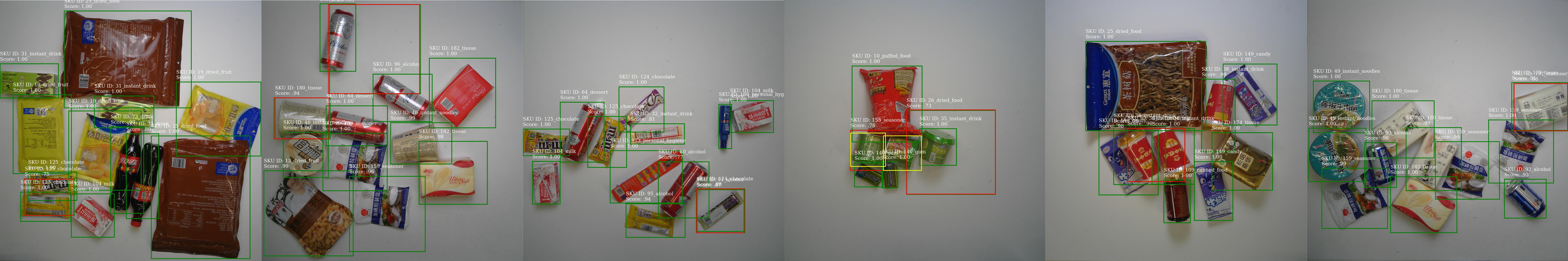} \label{fig:detfp} } \\

 \subfloat[\textbf{Successful cases for different clutter modes.} From left to right: \texttt{easy}, \texttt{medium}, and \texttt{hard} modes. The green bounding boxes are the corrected predictions with the product IDs and the confidence scores.] { \includegraphics[width=0.92\textwidth]{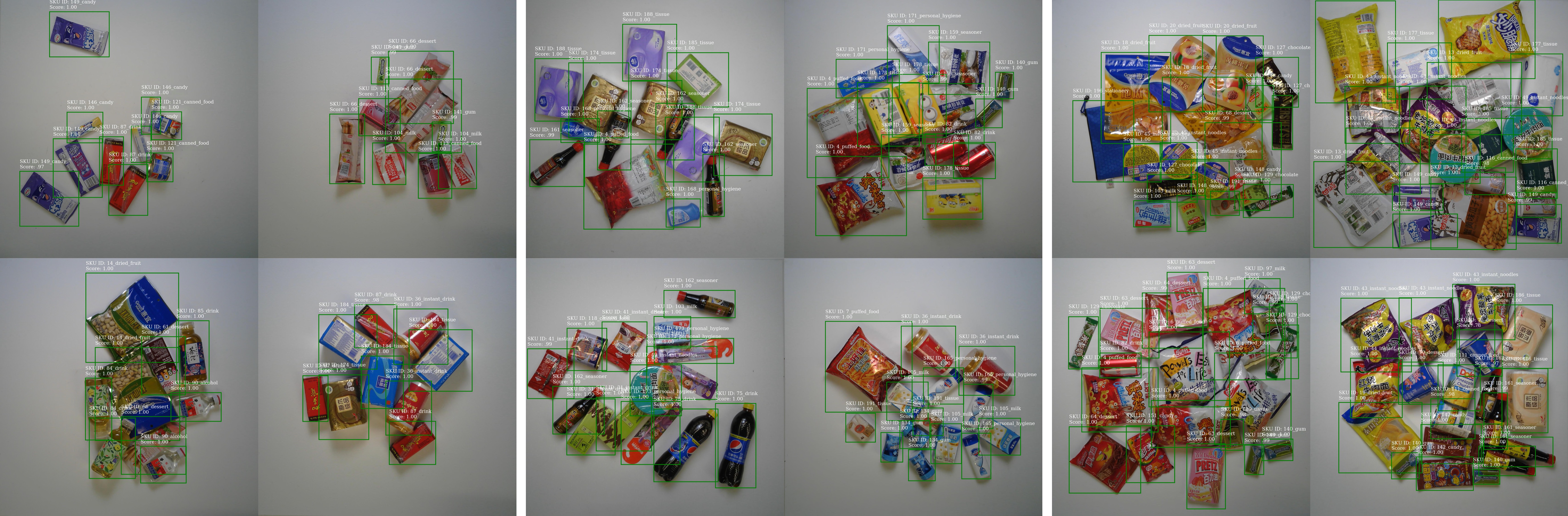} \label{fig:dethardbutright} }
 
\vspace{0.5em}
 \caption{Failure and successful cases of the ACO task on the proposed RPC dataset. These detection results are obtained by the \textbf{Syn+Render} method trained on 200,000 training images.} \label{fig:detresults}
\end{figure*}

\clearpage
\section{The ACO performance \vs the number of synthesized and rendered images via domain translation}

In this section, we study the affect of the number of training images (\ie, ``\textbf{Syn+Render}'') on the final checkout recognition performance. We change the number of training images in a set of $\left\{6250, 12500, 25000, 50000, 100000, 200000\right\}$, where $200000$ is the number of training images used in the experiments (cf. Table~3 in the paper). In each setting, the number of the synthesized checkout images equals to the number of the rendered checkout images via domain translation.

As shown, performance is boosted when the number of training images is increased, especially for the \texttt{Medium} and \texttt{Hard} modes. However, for \texttt{Easy}, the best performance is obtained when $100000$ training images are used.

\begin{table*}[h!]
    \caption{Main results of the chekcout task on our RPC dataset.} \label{table:results}
    \vspace{0.5em}
    \centering
    \small
    \begin{tabular}{|c|c|c|c|c|c|c|c|}
    \hline
        \textit{Clutter mode} & \textit{$\sharp$ synthesized \& rendered images} & \textit{cAcc} ($\uparrow$) & \textit{ACD} ($\downarrow$) & \textit{mCCD} ($\downarrow$) & \textit{mCIoU} ($\uparrow$) & \textit{mAP50} ($\uparrow$) & \textit{mmAP} ($\uparrow$) \\
	\hline
	\hline
	\multirow{6}{*}{\texttt{Easy}} &  \quad 6,250  &  55.61\%  &  0.93  &  0.14  &  88.23\%  &  94.22\%  &  73.88\%  \\
		&  ~~12,500  &  60.54\%  &  0.80  &  0.12  &  89.85\%  &  95.04\%  &  74.78\%  \\
		&  ~~25,000  &  67.52\%  &  0.60  &  0.09  &  92.47\%  &  96.37\%  &  76.37\%  \\
		&  ~~50,000  &  70.36\%  &  0.52  &  \textbf{0.07}  &  93.32\%  &  96.93\%  &  77.86\%  \\
		&  100,000  &  71.11\%  &  0.53  &  0.08  &  93.19\%  &  97.23\%  &  78.67\%  \\
		&  200,000  &  \textbf{73.17\%}  &  \textbf{0.49}  &  \textbf{0.07}  &  \textbf{93.66\%}  &  \textbf{97.34\%}  &  \textbf{79.01\%}  \\
	\hline
		\multirow{6}{*}{\texttt{Medium}}  &  \quad 6,250  &  32.98\%  &  1.76  &  0.15  &  86.33\%  &  91.79\%  &  66.61\%  \\
		&  ~~12,500  &  38.54\%  &  1.46  &  0.13  &  88.82\%  &  93.21\%  &  68.02\%  \\
		&  ~~25,000  &  47.20\%  &  1.10  &  0.10  &  91.37\%  &  94.91\%  &  70.02\%  \\
		&  ~~50,000  &  51.98\%  &  0.98  &  \textbf{0.08}  &  92.41\%  &  95.77\%  &  71.77\%  \\
		&  100,000  &  53.95\%  &  \textbf{0.90}  &  \textbf{0.08}  &  \textbf{92.98\%}  &  96.54\%  &  72.92\%  \\
		&  200,000  &  \textbf{54.69\%}  &  \textbf{0.90}  &  \textbf{0.08}  &  {92.95\%}  &  \textbf{96.56\%}  &  \textbf{73.24\%}  \\
	\hline
		\multirow{6}{*}{\texttt{Hard}}  &  \quad 6,250  &  19.06\%  &  2.64  &  0.15  &  85.79\%  &  90.77\%  &  64.78\%  \\
		&  ~~12,500  &  24.94\%  &  2.18  &  0.13  &  88.46\%  &  92.59\%  &  66.49\%  \\
		&  ~~25,000  &  34.65\%  &  1.58  &  0.09  &  91.37\%  &  94.37\%  &  68.99\%  \\
		&  ~~50,000  &  40.31\%  &  1.36  &  0.08  &  92.55\%  &  95.33\%  &  70.83\%  \\
		&  100,000  &  41.14\%  &  1.32  &  0.08  &  92.87\%  &  96.18\%  &  72.13\%  \\
		&  200,000  &  \textbf{42.48\%}  &  \textbf{1.28}  &  \textbf{0.07}  &  \textbf{93.06\%}  &  \textbf{96.45\%}  &  \textbf{72.72\%} \\
        \hline
	\hline
        \multirow{6}{*}{{Averaged}} & \quad 6,250 & 35.85\%  &  1.78  &  0.15  &  86.53\%  &  91.73\%  &  66.78\% \\
         & ~~12,500 & 41.23\%  &  1.47  &  0.12  &  88.99\%  &  93.24\%  &  68.25\% \\
         & ~~25,000 & 49.68\%  &  1.09  &  0.09  &  91.62\%  &  94.81\%  &  70.38\% \\
         & ~~50,000 & 54.05\%  &  0.95  &  0.08  &  92.72\%  &  95.69\%  &  72.16\%\\
         & 100,000 & 55.26\%  &  0.92  &  0.08  &  93.02\%  &  96.42\%  &  73.39\% \\
         & 200,000 & \textbf{56.68\%}  &  \textbf{0.89}  &  \textbf{0.07}  &  \textbf{93.19\%}  &  \textbf{96.57\%}  &  \textbf{73.83\%} \\
    \hline
\end{tabular}
\end{table*}

\clearpage
\section{Detailed statistical ACO results on single meta-category and single product category}

\begin{figure*}[h!]
	\centering
	\includegraphics[width=0.5\textwidth]{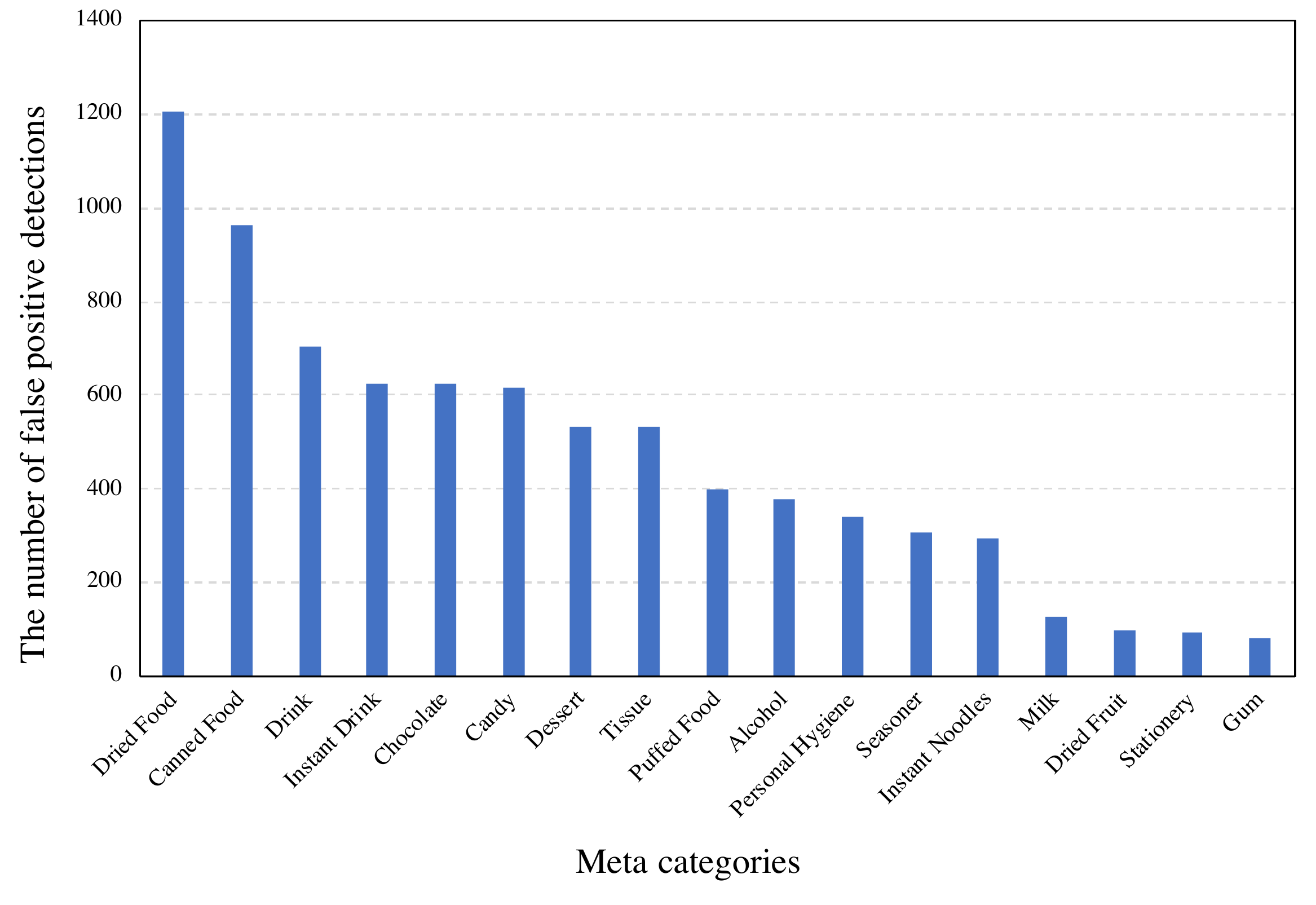} 
	\vspace{-0.5em}
	\caption{The number of false positive detection on each meta-category.}
	\label{fig:meta_fpnum}
\end{figure*}

\begin{figure*}[h!]
	\centering
	\includegraphics[width=0.5\textwidth]{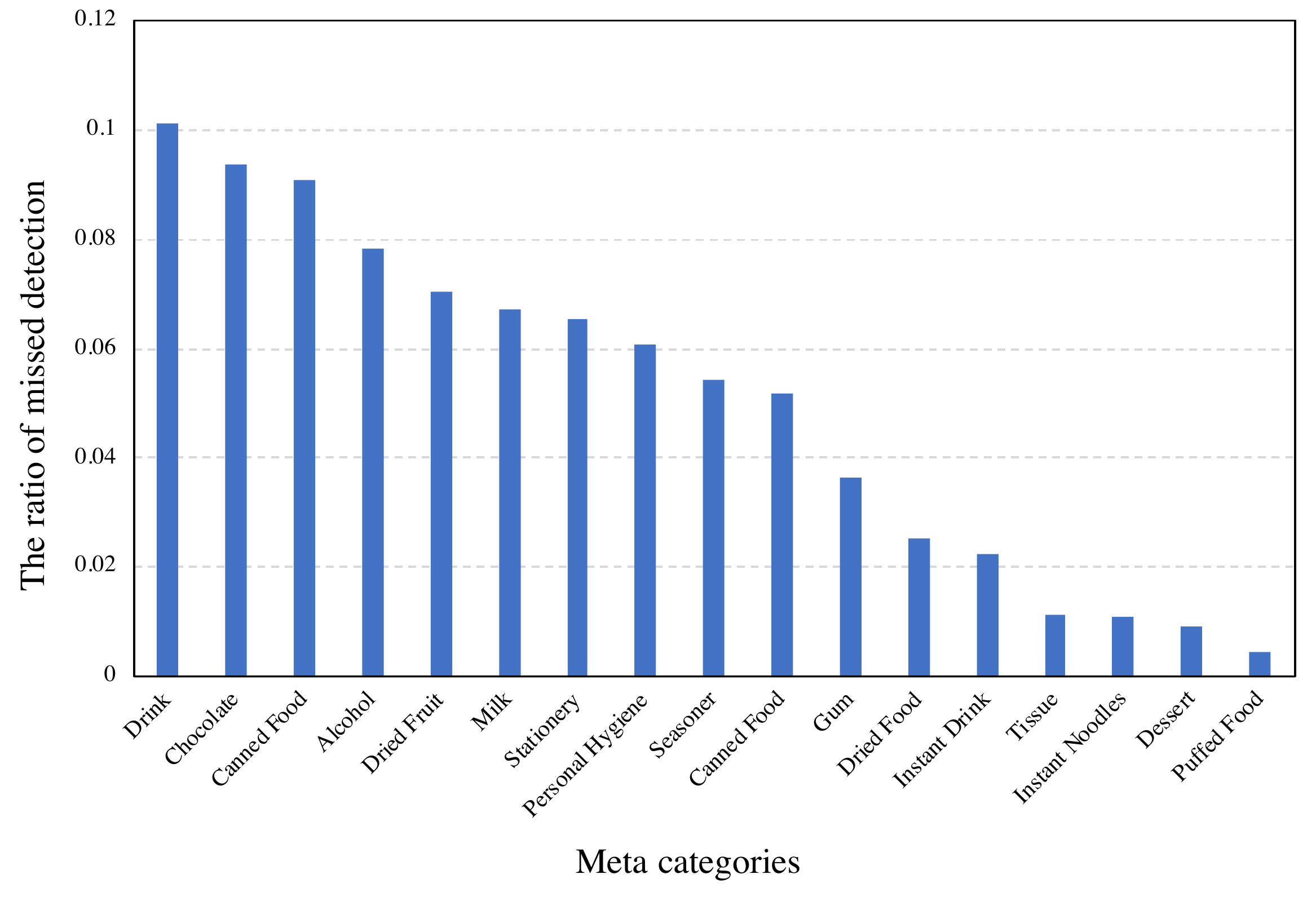} 
	\vspace{-0.5em}
	\caption{The ratio of missed detection on each meta-category. Concretely, the ratio is obtained via the number of missed detection divided by the number of ground truth bounding boxes of each meta-category.}
	\label{fig:meta_missrate}
\end{figure*}

\begin{figure*}[h!]
	\centering
	\includegraphics[width=0.5\textwidth]{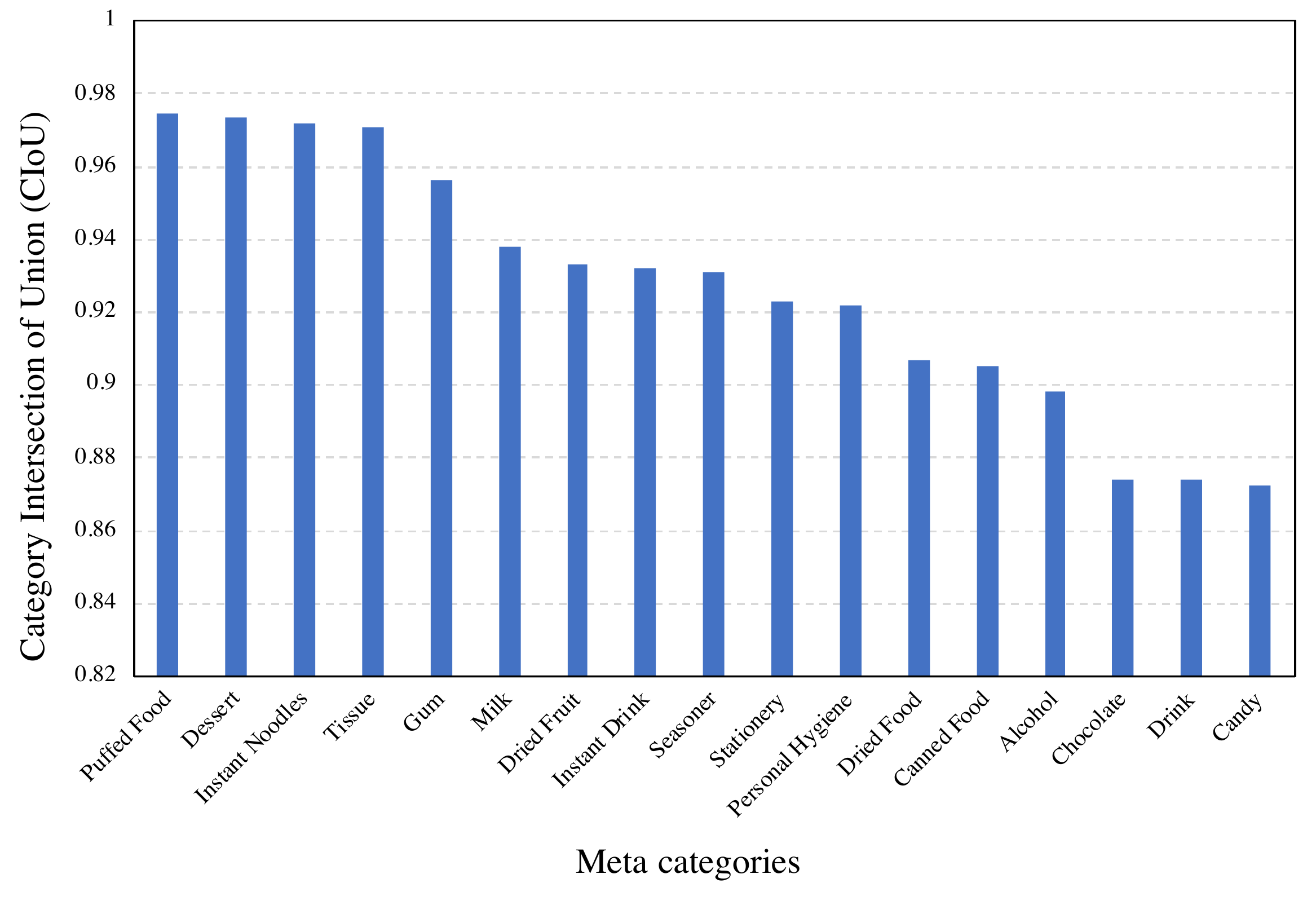} 
	\vspace{-0.5em}
	\caption{Performance of Category Intersection of Union (CIoU) on each meta-category, cf. Eq~(\ref{eq:mciou}).}
	\label{fig:meta_ciou}
\end{figure*}

\begin{figure*}[h]
	\centering
	\includegraphics[width=0.25\textwidth]{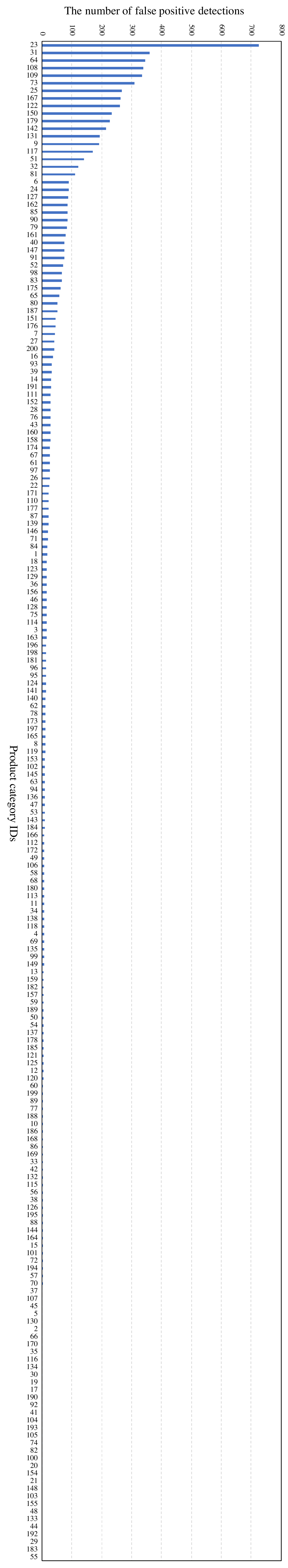}
	\caption{The number of false positive detection on each product category.}
	\label{fig:200_fpnum}
\end{figure*}

\begin{figure*}[h]
	\centering
	\includegraphics[width=0.25\textwidth]{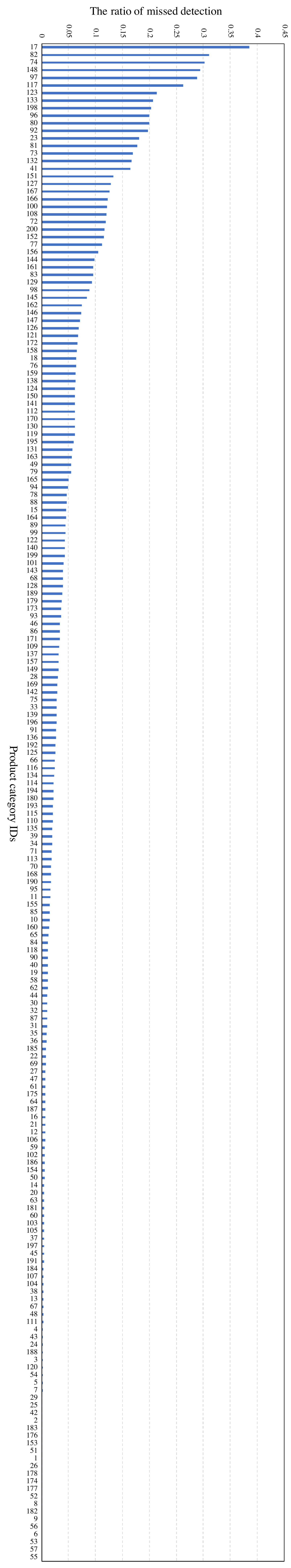}
	\caption{The ratio of missed detection on each product category. Concretely, the ratio is obtained via the number of missed detection divided by the number of ground truth bounding boxes of each product category.}
	\label{fig:200_missrate}
\end{figure*}

\begin{figure*}[h]
	\centering
	\includegraphics[width=0.25\textwidth]{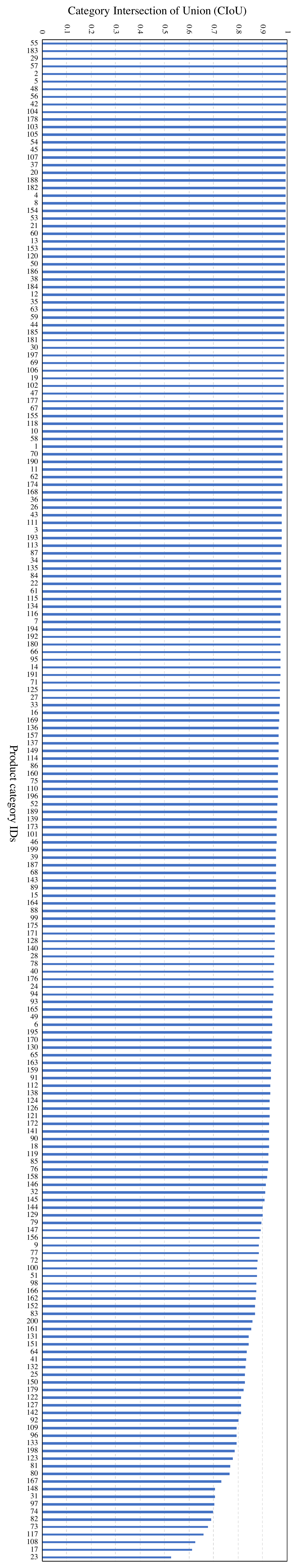}
	\caption{Performance of Category Intersection of Union (CIoU) on each product category, cf. Eq~(\ref{eq:mciou}).}
	\label{fig:200_ciou}
\end{figure*}

\clearpage
\twocolumn
{\small
\bibliographystyle{ieee}
\bibliography{checkoutdataset}

\begin{thebibliography}{10}\itemsep=-1pt

\bibitem{ocounting}
S.~Aich and I.~Stavness.
\newblock Improving object counting with heatmap regulation.
\newblock {\em CoRR}, abs/1803.05494, 2018.

\bibitem{NIPS2016_6399}
X.~Chen, Y.~Duan, R.~Houthooft, J.~Schulman, I.~Sutskever, and P.~Abbeel.
\newblock {InfoGAN}: Interpretable representation learning by information
  maximizing generative adversarial nets.
\newblock In {\em {Proc. Advances in Neural Inf. Process. Syst.}} 2016.

\bibitem{Cordts2016Cityscapes}
M.~Cordts, M.~Omran, S.~Ramos, T.~Rehfeld, M.~Enzweiler, R.~Benenson,
  U.~Franke, S.~Roth, and B.~Schiele.
\newblock The cityscapes dataset for semantic urban scene understanding.
\newblock In {\em {Proc. IEEE Conf. Comp. Vis. Patt. Recogn.}}, 2016.

\bibitem{pascal-voc-2012}
M.~Everingham, L.~Van~Gool, C.~K.~I. Williams, J.~Winn, and A.~Zisserman.
\newblock The {PASCAL} {V}isual {O}bject {C}lasses {C}hallenge 2012 {(VOC2012)}
  {R}esults.
\newblock
  \url{http://www.pascal-network.org/challenges/VOC/voc2012/workshop/index.html}.

\bibitem{ulrichmvtec}
P.~Follmann, T.~Bottger, P.~Hartinger, R.~Konig, and M.~Ulrich.
\newblock {MVTec D2S}: Densely segmented supermarket dataset.
\newblock In {\em {Proc. Eur. Conf. Comp. Vis.}}, 2018.

\bibitem{da}
Y.~Ganin and V.~Lempitsky.
\newblock Unsupervised domain adaptation by backpropagation.
\newblock In {\em {Proc. Int. Conf. Mach. Learn.}}, 2015.

\bibitem{george2014recognizing}
M.~George and C.~Floerkemeier.
\newblock Recognizing products: A per-exemplar multi-label image classification
  approach.
\newblock In {\em {Proc. Eur. Conf. Comp. Vis.}}, 2014.

\bibitem{GANs}
I.~Goodfellow, J.~Pouget-Abadie, M.~Mirza, B.~Xu, D.~Warde-Farley, S.~Ozair,
  A.~Courville, and Y.~Bengio.
\newblock Generative adversarial nets.
\newblock In {\em {Proc. Advances in Neural Inf. Process. Syst.}} 2014.

\bibitem{He2016DeepRL}
K.~He, X.~Zhang, S.~Ren, and J.~Sun.
\newblock Deep residual learning for image recognition.
\newblock {\em {Proc. IEEE Conf. Comp. Vis. Patt. Recogn.}}, 2016.

\bibitem{hu2018cvprsenet}
J.~Hu, L.~Shen, S.~Albanie, G.~Sun, and E.~Wu.
\newblock Squeeze-and-excitation networks.
\newblock In {\em {Proc. IEEE Conf. Comp. Vis. Patt. Recogn.}}, 2018.

\bibitem{ping2016detectingTIP}
P.~Hu, W.~Wang, C.~Zhang, and K.~Lu.
\newblock Detecting salient objects via color and texture compactness
  hypotheses.
\newblock {\em {{IEEE} Trans. Image Process.}}, pages 4653--4664, 2016.

\bibitem{Huang_2018_ECCV}
S.-W. Huang, C.-T. Lin, S.-P. Chen, Y.-Y. Wu, P.-H. Hsu, and S.-H. Lai.
\newblock {AugGAN}: Cross domain adaptation with {GAN}-based data augmentation.
\newblock In {\em {Proc. Eur. Conf. Comp. Vis.}}, 2018.

\bibitem{jund2016freiburg}
P.~Jund, N.~Abdo, A.~Eitel, and W.~Burgard.
\newblock The freiburg groceries dataset.
\newblock {\em arXiv preprint arXiv:1611.05799}, 2016.

\bibitem{adam}
D.~P. Kingma and J.~Ba.
\newblock Adam: {A} method for stochastic optimization.
\newblock {\em CoRR}, abs/1412.6980, 2014.

\bibitem{journals/corr/KingmaW13}
D.~P. Kingma and M.~Welling.
\newblock Auto-encoding variational bayes.
\newblock {\em CoRR}, abs/1312.6114, 2013.

\bibitem{koubaroulis2002evaluating}
D.~Koubaroulis, J.~Matas, J.~Kittler, and C.~CMP.
\newblock Evaluating colour-based object recognition algorithms using the
  {SOIL}-47 database.
\newblock In {\em {Proc. Asian Conf. Comp. Vis.}}, 2002.

\bibitem{krahenbuhl2011efficient}
P.~Kr{\"a}henb{\"u}hl and V.~Koltun.
\newblock Efficient inference in fully connected {CRF}s with {G}aussian edge
  potentials.
\newblock In {\em {Proc. Advances in Neural Inf. Process. Syst.}}, 2011.

\bibitem{NIPS2012_4824}
A.~Krizhevsky, I.~Sutskever, and G.~E. Hinton.
\newblock Image{N}et classification with deep convolutional neural networks.
\newblock In {\em {Proc. Advances in Neural Inf. Process. Syst.}} 2012.

\bibitem{5206594}
C.~H. Lampert, H.~Nickisch, and S.~Harmeling.
\newblock Learning to detect unseen object classes by between-class attribute
  transfer.
\newblock In {\em {Proc. IEEE Conf. Comp. Vis. Patt. Recogn.}}, 2009.

\bibitem{lin2017feature}
T.-Y. Lin, P.~Doll{\'a}r, R.~Girshick, K.~He, B.~Hariharan, and S.~Belongie.
\newblock Feature pyramid networks for object detection.
\newblock In {\em {Proc. IEEE Conf. Comp. Vis. Patt. Recogn.}}, 2017.

\bibitem{mscoco}
T.-Y. Lin, M.~Maire, S.~Belongie, J.~Hays, P.~Perona, D.~Ramanan,
  P.~Doll{\'a}r, and C.~L. Zitnick.
\newblock Microsoft {COCO}: Common objects in context.
\newblock In {\em {Proc. Eur. Conf. Comp. Vis.}}, 2014.

\bibitem{least2017mao}
X.~Mao, Q.~Li, H.~Xie, R.~Y. Lau, Z.~Wang, and S.~P. Smolley.
\newblock Least squares generative adversarial networks.
\newblock In {\em {Proc. IEEE Conf. Comp. Vis. Patt. Recogn.}}, 2017.

\bibitem{merler2007recognizing}
M.~Merler, C.~Galleguillos, and S.~Belongie.
\newblock Recognizing groceries in situ using in vitro training data.
\newblock In {\em {Proc. IEEE Conf. Comp. Vis. Patt. Recogn.}}, 2007.

\bibitem{DBLP:journals/corr/abs-1709-00663}
A.~Mishra, M.~S.~K. Reddy, A.~Mittal, and H.~A. Murthy.
\newblock A generative model for zero shot learning using conditional
  variational autoencoders.
\newblock {\em CoRR}, abs/1709.00663, 2017.

\bibitem{yolo2017redmon}
J.~Redmon and A.~Farhadi.
\newblock {YOLO9000: Better, faster, stronger}.
\newblock In {\em {Proc. IEEE Conf. Comp. Vis. Patt. Recogn.}}, 2017.

\bibitem{rocha2010automatic}
A.~Rocha, D.~C. Hauagge, J.~Wainer, and S.~Goldenstein.
\newblock Automatic fruit and vegetable classification from images.
\newblock {\em Computers and Electronics in Agriculture}, pages 96--104, 2010.

\bibitem{Simonyan14c}
K.~Simonyan and A.~Zisserman.
\newblock Very deep convolutional networks for large-scale image recognition.
\newblock {\em CoRR}, abs/1409.1556, 2014.

\bibitem{tobin2017domain}
J.~Tobin, R.~Fong, A.~Ray, J.~Schneider, W.~Zaremba, and P.~Abbeel.
\newblock Domain randomization for transferring deep neural networks from
  simulation to the real world.
\newblock In {\em IROS}, 2017.

\bibitem{wang2018gcnzeroshot}
X.~Wang, Y.~Ye, and A.~Gupta.
\newblock Zero-shot recognition via semantic embeddings and knowledge graphs.
\newblock In {\em {Proc. IEEE Conf. Comp. Vis. Patt. Recogn.}}, 2018.

\bibitem{8237572}
Z.~Yi, H.~Zhang, P.~Tan, and M.~Gong.
\newblock Dualgan: Unsupervised dual learning for image-to-image translation.
\newblock In {\em {Proc. IEEE Int. Conf. Comp. Vis.}}, 2018.

\bibitem{Zhou_2018_CVPR}
Y.~Zhou and L.~Shao.
\newblock Viewpoint-aware attentive multi-view inference for vehicle
  re-identification.
\newblock In {\em {Proc. IEEE Conf. Comp. Vis. Patt. Recogn.}}, 2018.

\bibitem{zhu2017unpaired}
J.-Y. Zhu, T.~Park, P.~Isola, and A.~A. Efros.
\newblock Unpaired image-to-image translation using cycle-consistent
  adversarial networks.
\newblock In {\em {Proc. IEEE Int. Conf. Comp. Vis.}}, 2017.

\end{thebibliography}
}

\end{document}